%% file: acl_final.tex
\PassOptionsToPackage{dvipsnames}{xcolor}

\documentclass[11pt]{article}

\usepackage[preprint]{acl}

\usepackage{times}
\usepackage{latexsym}

\usepackage[T1]{fontenc}

\usepackage[utf8]{inputenc}

\usepackage{microtype}

\usepackage{inconsolata}

\usepackage{graphicx}

\usepackage{pifont}
\usepackage{algorithm}
\usepackage{algpseudocode}
\usepackage{array}
\usepackage{multirow}
\usepackage{makecell}
\usepackage{booktabs}
\usepackage{bbding}
\usepackage{colortbl}
\usepackage{subcaption}
\usepackage{placeins} 

\usepackage{times}
\usepackage{latexsym}

\usepackage[T1]{fontenc}

\usepackage[utf8]{inputenc}

\usepackage{microtype}

\usepackage{inconsolata}

\usepackage{xspace}
\usepackage{adjustbox}
\usepackage{amssymb}
\usepackage{longtable}
\usepackage{enumitem}

\usepackage[most]{tcolorbox}
\definecolor{myGray}{RGB}{64, 64, 64}
\definecolor{myLightGray}{RGB}{242, 242, 242}

\newtcolorbox{prompt}[2][]{
    colback=myLightGray,
    colframe=myGray,
    fonttitle=\bfseries\small,
    boxrule=0.4mm,
    fontupper=\small, 
    fontlower=\small,
    coltitle=white,
    title=#2,
    #1,breakable
}
\usepackage{cleveref}
\usepackage{arydshln}

\definecolor{myPurple}{RGB}{204, 0, 204}
\newcommand{\cmark}{\textcolor{ForestGreen}{\ding{51}}}
\newcommand{\xmark}{\textcolor{red}{\ding{55}}}
\newcommand{\wmark}{\raisebox{-2pt}{\includegraphics[width=1em]{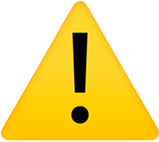}}}
\newcommand\mirror{\raisebox{-2.5pt}{\includegraphics[width=0.7em]{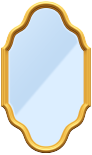}\space}}

\definecolor{warningcolor}{RGB}{250,35,64}
\title{\mirror\textsc{Mirror}: Multimodal Cognitive Reframing Therapy\\for Rolling with Resistance
}

\author{
  Subin Kim$^{1}$\thanks{~Equal contribution}\thanks{~This work was done while at POSTECH.},  \textbf{Hoonrae Kim}$^{2}$\footnotemark[1], 
  \textbf{Jihyun Lee}$^{2}$\footnotemark[1],  
  \textbf{Yejin Jeon}$^{2}$\footnotemark[1],  
  \textbf{Gary Geunbae Lee}$^{2, 3}$ \quad \\
  KT Corporation, Republic of Korea$^{1}$ \\
Graduate School of Artificial Intelligence, POSTECH, Republic of Korea$^{2}$ \\
Computer Science and Engineering, POSTECH, Republic of Korea$^{3}$ \\
 \normalsize \texttt{subin.k@kt.com}, 
  \texttt{\{hoonrae, jihyunlee, jeonyj0612, gblee\}@postech.ac.kr}\\
}

\begin{document}
\maketitle

\begin{abstract}

Recent studies have explored the use of large language models (LLMs) in psychotherapy; however, text-based cognitive behavioral therapy (CBT) models often struggle with client resistance, which can weaken therapeutic alliance. 
To address this, we propose a multimodal approach that incorporates nonverbal cues, which allows the AI therapist to better align its responses with the client's negative emotional state.
Specifically, we introduce a new synthetic dataset, \textsc{Mirror} (\textbf{M}ultimodal \textbf{I}nte\textbf{r}active \textbf{Ro}lling with \textbf{R}esistance), which is a novel synthetic dataset that pairs each client's statements with corresponding facial images. 
Using this dataset, we train baseline vision language models (VLMs) so that they can analyze facial cues, infer emotions, and generate empathetic responses to effectively manage client resistance.
These models are then evaluated in terms of both their counseling skills as a therapist, and the strength of therapeutic alliance in the presence of client resistance. 
Our results demonstrate that \textsc{Mirror} significantly enhances the AI therapist’s ability to handle resistance, which outperforms existing text-based CBT approaches.
Human expert evaluations further confirm the effectiveness of our approach in managing client resistance and fostering therapeutic alliance.

\end{abstract}

\section{Introduction}
\textbf{\textcolor{warningcolor}{Important:}} \textcolor{red}{We explore how vision-language models support digital CBT, but they should NOT replace professional psychological treatment.}

Cognitive reframing is a central part of cognitive behavioral therapy (CBT), which helps individuals replace negative and intrusive thoughts with more rational and balanced ones.
Towards this objective, large language models (LLMs) have recently shown great promise and are increasingly being explored in psychotherapy \cite{ziems-etal-2022-inducing, maddela-etal-2023-training, sharma-etal-2023-cognitive, qu2023conditioning, yang2023towards, 10.1145/3589334.3648137,  xiao-etal-2024-healme, na-2024-cbt, lee-etal-2024-cactus}. 
As such, these systems have actually been utilized in real-world applications as effective adjunct tools in psychotherapy, providing meaningful support for individuals with mental disorders such as depression and anxiety \cite{chatbot_helps1, chatbot_helps2, chatbot_helps3}\footnote{A comprehensive review of related work is provided in Appendix~\ref{sec:related_work}.}.

\input{figures/problem}

Despite this progress, the existing text-based CBT model struggles to detect and respond to client resistance \cite{wang2025evaluating}, which is a common therapeutic challenge that involves the client's reluctance or opposition to change. 
This resistance often stems from the directive nature of CBT, where structured interventions may unintentionally provoke discomfort or defensiveness \cite{patterson1994functional, Moyers2006-yy, Constantino2017-vk, westra2018using, hara2020comparing}. 
Left unaddressed, resistance can diminish therapeutic alliance and reduce treatment efficacy. 
It is crucial to note that such resistance is frequently conveyed through nonverbal cues like facial expressions, sighs, or posture shifts.
Due to this property, pure-text-based models fail to perceive resistance, which leads to premature advice-giving rather than addressing deeper emotional needs (Figure~\ref{fig:problem}). 
Addressing this limitation thus requires multimodal integration. 
However, collecting real multimodal psychotherapy data to train models to identify such multimodal cues, introduces severe privacy risks as sessions often involve deeply personal disclosures, including trauma, mental illness, and other confidential experiences.

\input{figures/dataset}

In this work, we propose a multimodal approach to cognitive reframing that integrates both textual and nonverbal information to better detect and manage client resistance. 
We introduce \textsc{Mirror} (\textbf{M}ultimodal \textbf{I}nte\textbf{r}active \textbf{Ro}lling with \textbf{R}esistance), which is a synthetic dataset designed to simulate real therapeutic interactions. 
Specifically, \textsc{Mirror} features generated dialogues between clients and therapists, annotated with client facial expressions reflecting three distinct types of resistance. 
We leverage LLMs to generate realistic session content, synthesize corresponding facial cues, and apply rigorous filtering to ensure quality and safety. 
This dataset enables the development of vision-language models (VLMs) tailored to CBT scenarios, where emotional alignment and alliance are essential.
In addition, we introduce \textit{emotional captioning}, an adaptation of chain-of-thought (CoT) prompting to the multimodal setting.
This strategy explicitly interprets the client’s emotional state through intermediate captions, which in turn guide the model toward generating emotionally attuned and context-aware responses in therapeutic dialogue.

We evaluate our approach using a VLM that is trained on the \textsc{Mirror} dataset and enhanced with \textit{planning} and \textit{emotional captioning}. 
Compared to existing LLMs and VLMs, our model demonstrates superior performance across therapist skill assessment, alliance building, and applicability to real counseling scenarios. 
The results highlight the importance of multimodal approaches in managing client resistance and improving CBT outcomes.

Our contributions are summarized as follows:
\begin{itemize}
    \item  We explore a multimodal cognitive reframing for coping with client resistance, and present \textsc{Mirror}, which features turn-level client facial expressions across diverse resistance types.
    \item We establish baseline models on the \textsc{Mirror} dataset and propose an \textit{emotional captioning} method, which helps VLMs generate emotionally aligned, vision-aware therapeutic responses.
\end{itemize}
To further support research in this area, we publicly release our code and dataset\footnote{\url{https://github.com/nobel-postech/mirror}}.

\section{Problem Definition}
\label{sec:problem_definition}
Our goal is to enhance the AI therapist’s ability to manage client resistance by integrating both verbal and nonverbal cues through a multimodal approach. 
To guide the development and evaluation of such models, we define two key assessment dimensions that reflect essential aspects of effective therapy:

\begin{itemize} 
    \item \textbf{Therapist Skills Assessment:} Evaluates the AI therapist's competence in two key categories of general counseling skills and CBT-specific techniques. 

    \item \textbf{Client Alliance Assessment:} Focuses on the AI therapist's ability to establish a strong therapeutic bond, which is critical for reducing resistance and promoting positive outcomes.
\end{itemize}

\section{\mirror\textsc{Mirror}: Multimodal Interactive Rolling with Resistance Dataset}
As illustrated in Figure \ref{fig:dataset}, the \textsc{Mirror} dataset is constructed through three main steps, which is followed by a comprehensive quality and safety validation process.
Through dataset synthesis, we generate over 3,000 multimodal counseling dialogues, with each client turn annotated with a facial expression image that captures the client’s emotional state\footnote{All used prompts are provided in Appendix \ref{sec:mirror_prompts}.}.

\subsection{Step 1: Multimodal Dialogue Design}
\label{sec:step1}

To build the multimodal dialogue design for \textsc{Mirror}, we combine facial and textual data from two sources: CelebA \cite{liu2015faceattributes} for facial expressions and \textsc{Cactus} \cite{lee-etal-2024-cactus} for text-based cognitive reframing therapy. 
While \textsc{Cactus} is originally a text-only dialogue dataset, we only extract its underlying structured profiles, which includes client intake forms, thinking traps, counseling plans, and CBT techniques. 

In order to assign a facial identity to each client, we pair every \textsc{Cactus} profile with a CelebA image based on gender and age predictions from the DeepFace library \cite{serengil2021lightface}. 

We further augment each client profile with four distinct resistance types: cognitive, emotional, behavioral, and non-resistant, following the taxonomy proposed by \citet{beal2013effect}. 
Rather than assigning a single resistance label to each profile, we generate four variants per client, each conditioned on a different resistance type. 
This results in four variants per client, allowing the model to encounter diverse resistance behaviors in the same therapeutic context and ensuring class balance across the dataset.
This process yields a complete multimodal dialogue setup for each client, where a structured \textsc{Cactus} profile, a facial identity, and a specified resistance type are jointly configured.
The resulting design supports therapeutically grounded dialogue generation based on client context and CBT plan.

\input{tables/dataset_comparison}

\subsection{Step 2: Counseling Screenplay Generation}
\label{sec:step2}

We synthesize counseling dialogues in the form of screenplays rather than plain transcripts, to more naturally reflect the emotional nuance of real therapeutic interactions.
A key advantage of this format is its explicit representation of nonverbal cues through stage directions (e.g., \textit{[slightly defensive, arms crossed]'} in  Figure~\ref{fig:dataset}). 

These stage directions serve two critical purposes:  
(1) They enrich the textual context by capturing subtle emotional dynamics that are characteristic of real therapy sessions.  
(2) They act as structured signals for downstream facial expression synthesis, which ensures the generation of consistent and emotionally aligned client images.
Based on the predefined profiles, these screenplays are generated using \textsc{GPT-4o-mini}\footnote{Version {\fontfamily{qcr}\selectfont gpt-4o-mini-2024-07-18}.}.

\subsection{Step 3: Facial Expression Synthesis}
\label{sec:step3}

After constructing the screenplay, we synthesize turn-level facial expressions that reflect the emotional dynamics conveyed through both verbal content and stage directions. The key contribution of this step lies in designing a prompt construction method that encodes nonverbal cues into the image generation process.

We leverage \textbf{PhotoMaker}~\cite{li2023photomaker}, which is a diffusion-based model that takes three inputs: a reference image to preserve facial identity, a positive prompt for the desired expression, and a negative prompt to suppress conflicting features. 
To generate these prompts, we condition \textsc{LLaMA-3-8B}~\cite{llama3modelcard} on the full client utterance, which includes inline stage directions (see Figure~\ref{fig:dataset}).
As a result, \textsc{LLaMA-3-8B} produces two facial expression descriptions: a target expression (e.g., “downcast expression with eyes looking away”) and a contrasting one (e.g., “trusting expression with a gentle smile”), which populate the positive and negative prompts, respectively.

This approach enables the synthesis of emotionally aligned client images throughout the dialogue. 
As shown in Figure~\ref{fig:dataset}, expressions like \textit{[looking away]} are clearly expressed in the synthesized images.
By translating nonverbal cues into structured prompts, we ensure that facial expressions reflect the client’s emotional state, even when the textual utterance alone does not explicitly convey it.
The role of stage direction in image synthesis is further examined in Appendix~\ref{sec:stage_direction}.

\subsection{Step 4: Filtering for Quality and Safety}
\label{sec:step4}

\paragraph{Dataset Quality Filtering}
To ensure the overall quality and coherence with image of multimodal counseling dialogues, we apply six filtering approaches: 

\textbf{\textit{(1) Image-Text Similarity Filtering}} uses CLIP \cite{pmlr-v139-radford21a}, following prior use in vision–language filtering \cite{Howard_2024_CVPR}, to measure alignment between generated images and stage directions; cases with low similarity (below 0.2) are discarded (2.95\% rejected).  
\textbf{\textit{(2) Identity Preservation Filtering}} employs ArcFace \cite{deng2019arcface} to maintain facial similarity across dialogue turns, adopting the approach of \citet{Melzi_2023_ICCV} (66.05\% rejected\footnote{
The high rejection rate stems from our conservative threshold combined with the sensitivity of facial embeddings. 
In particular, cropped or partially occluded faces (e.g., head turns or partial coverage) and dynamic expressions often received low similarity scores, even when identities appeared visually consistent. 
This reflects the strictness of our filtering criteria rather than instability in the generation model.
}).
\textbf{\textit{(3) Gender Preservation Filtering}} also follows \citet{Melzi_2023_ICCV}, using DeepFace to ensure that the detected gender matches the client’s multimodal profile (15.39\% rejected).  
\textbf{\textit{(4) Basic Filtering}} eliminates dialogues that contain utterances longer than 100 words or too few (fewer than 4) or too many (more than 20) conversation turns, following practices in large-scale dialogue datasets \cite{kim-etal-2023-soda, lee-etal-2024-stark}. 
In addition, we applied a custom rule to filter out utterances that exhibit unnatural repetition of the same part-of-speech more than three times in a row (Overall, 1.03\% rejected).  
\textbf{\textit{(5) Copy-Paste Filtering}} removes instances where client personas are unnaturally stated instead of contextually integrated, following \citet{lee-etal-2022-personachatgen} (1.36\% rejected).  
\textbf{\textit{(6) Therapeutic Alliance Filtering}} assesses the quality of the counseling interactions using \textsc{GPT-4o}\footnote{Version {\fontfamily{qcr}\selectfont gpt-4o-2024-08-06}.} to evaluate WAI\footnote{WAI stands for Working Alliance Inventory.} \cite{li-etal-2024-understanding-therapeutic}. 
While prior work did not use WAI for filtering, they reported moderate agreement between GPT-based and human expert ratings (ICC $\approx$ 0.66--0.72), supporting the reliability of this approach. In our dataset, dialogues with an average WAI score below 0.3 were discarded (10.01\% rejected).

\paragraph{Dataset Safety Filtering}
To uphold ethical standards and prevent harmful content, we apply two additional approaches. 
\textbf{\textit{(1) NSFW Filtering}} uses a Not-Safe-For-Work (NSFW) detector\footnote{\url{https://huggingface.co/Falconsai/nsfw_image_detection}} to remove images that are visually unsuitable for mental health dialogue contexts.
\textbf{\textit{(2) Dialogue Safety Filtering}} leverages Canary \cite{kim-etal-2022-prosocialdialog} to identify and eliminate instances containing toxic, unethical, or unsafe language, in accordance with prior safety protocols~\cite{kim-etal-2023-soda, lee-etal-2024-stark} (1.09\% rejected).
These layered filtering stages are critical for constructing a high-quality dataset that is not only realistic and coherent but also ethically robust and clinically applicable.


\subsection{Comparative Analysis of \textsc{Mirror}}

Through the preceding stages, we have curated the first multimodal CBT dataset that explicitly incorporates client resistance. As shown in Table~\ref{tab:dataset_comparison}, \textsc{Mirror} contains a comparatively large number of dialogues with high turn density and dynamic visual responses. 
Unlike prior datasets such as M2CoSC \cite{kim-etal-2025-multimodal}, which uses a single static image per dialogue, or MEDIC \cite{10.1145/3581783.3612346}, which is limited to a single turn, \textsc{Mirror} provides image sequences that evolve turn-by-turn in alignment with client emotion.

\section{Reasoning Strategies}
\label{sec:reasoning_strategies}
To analyze how structured reasoning can affect resistance management, we explore two strategies: \textit{planning} and \textit{emotional captioning}. 
These strategies provide useful insights into how pre-session reasoning and multimodal inputs may shape AI therapists’ responses.

\paragraph{Planning}
Following \citet{lee-etal-2024-cactus}, we adopt a pre-session \textit{planning} step in which the model infers a counseling strategy based on the client’s profile (e.g., name, age, gender, occupation) and counseling objectives. 
This inferred plan is then used to guide the model’s responses during the session (Figure~\ref{fig:planning}). 
The approach is intended to help the AI therapist maintain a facilitative role and select appropriate CBT techniques, rather than directly correcting client statements. 

\input{figures/planning}

\paragraph{Emotional Captioning}
To handle client resistance more effectively, we also incorporate \textit{emotional captioning}, a reasoning module that interprets the client’s emotional state from facial expressions.
At each dialogue turn, the model receives a facial image and generates a short textual description of the client’s emotional state (e.g., \textit{looking down, slightly defensive}), which is then used to guide the AI therapist’s response (Figure~\ref{fig:emotional_captioning}). 
By grounding the model’s behavior in visual cues, \textit{emotional captioning} supplements verbal input with nonverbal affective signals, improving alignment with the client's psychological state\footnote{Prompt templates used in this process are detailed in Appendix~\ref{sec:counseling_prompts}.}.

\input{figures/emotional_caption}


\section{Experimental Settings}
\label{sec:experimental_settings}

Following \citet{smith-etal-2022-human, liu2023chatcounselor}, and \citet{lee-etal-2024-cactus}, we assess the AI therapist based on full simulated counseling sessions rather than turn-level assessments. 
Each session involves an AI therapist interacting with a virtual client exhibiting varying types of resistance. 
We compare different model variants to assess the contribution of \textit{planning} and \textit{emotional captioning} strategies.

\input{tables/gpt_counselingeval}

\subsection{Client Agents with Resistance}

We adopt \textsc{GPT-3.5-Turbo}\footnote{Version {\fontfamily{qcr}\selectfont gpt-3.5-turbo-0125}.} as the virtual client and conduct simulations based on predefined multimodal profiles. 
Unlike the training data, which contains fully generated multimodal dialogues (3,073 in total), the evaluation data are independent of \textsc{Mirror} and consist of 800 newly constructed client profiles generated using the same procedure but without associated dialogues.
These profiles include three resistant types, with 200 examples for each. 
In addition, 200 non-resistant cases were included for comparison to highlight the performance drop when models face resistant clients. 
All reported evaluation results focus on the three resistant types, with non-resistant cases used only for relative comparison.

Each session is considered terminated if the client attempts to disengage after two consecutive turns.

Within each client's utterance, nonverbal cues are embedded as stage directions within brackets, as described in \S\ref{sec:step3}. Note that these cues are used for facial expression generation and are invisible to the AI therapist when generating responses.
For facial expressions, we generate LLM-based client's images at each turn, following the same process used in dataset construction (\S\ref{sec:step3} and \S\ref{sec:step4}). 
Appendix \ref{sec:counseling_prompts} provides client setup and simulation prompt details.

\subsection{AI Therapist Model Variations and Baselines}

\paragraph{AI Therapist Baselines}
Our primary baseline, \textsc{Mirror-LLaVA}, is a \textsc{LLaVA-v1.5-7B}~\cite{llava} trained on the \textsc{Mirror} dataset.
To examine the benefit of multimodal integration, we include \textsc{Camel-LLaMA3}\footnote{\url{https://huggingface.co/cactus-camel/camel-llama3}}, a text-only CBT model trained on therapeutic dialogues~\cite{lee-etal-2024-cactus}.
We also evaluate general-purpose models that are not fine-tuned for counseling: \textsc{LLaMA-3-8B}, \textsc{LLaVA-v1.5-7B}, and \textsc{GPT-3.5-Turbo}. These serve as non-specialized baselines to assess the impact of domain adaptation and modality alignment.
Although evaluating against vision-capable GPT-4o would provide a stronger performance reference, budget and accessibility constraints prevented us from including it in this version.

Further implementation details, including training procedures, are provided in Appendix~\ref{sec:training}.

\paragraph{Reasoning Variants}
We evaluate two structured reasoning strategies introduced in Section~\ref{sec:reasoning_strategies}: 
\textit{planning}, a pre-session process that infers a counseling strategy from the client’s profile, 
and \textit{emotional captioning}, which generates short textual descriptions of the client’s facial expressions to guide therapist responses. 
For clarity, we denote models incorporating these strategies with the subscripts $P$ (\textit{planning}) and $EC$ (\textit{emotional captioning}).

\subsection{Metrics for Assessment}
As defined in \S\ref{sec:problem_definition}, we evaluate the therapist’s ability to manage client resistance across two key areas: therapist skills and client alliance.
The evaluation prompts are provided in Appendix~\ref{sec:evaluating_ai_therapist}.

\textbf{Therapist skills} are assessed using the \textsc{CounselingEval} framework \cite{lee-etal-2024-cactus}, which covers both general counseling skills and CBT-specific competencies. 
In particular, general counseling skills encompass the ability to interpret client concerns (Understanding), maintain a therapeutic relationship (Interpersonal Effectiveness), and facilitate collaborative decision-making (Collaboration).
Meanwhile, CBT-specific skills evaluate the ability to guide clients in discovering their thoughts (Guided Discovery) and identify mal-adaptive patterns (Focus). Each component of the therapist’s skills is rated on a scale from 0 to 6\footnote{For our experiments, we do not use a Strategy score, which assesses the coherence of intervention strategies, as it strongly correlates with the length of the AI therapist's responses (see Appendix \ref{sec:gpt_analysis}).}.

\textbf{Client alliance} is measured following \citet{li-etal-2024-understanding-therapeutic}, which assesses agreement of therapy objectives (Goal), engagement in counseling tasks (Approach), and the strength of emotional connection (Affective Bond), and is scored from 1 to 5.

\section{Results and Discussion}
\subsection{Therapist Skills Assessment}
 Table \ref{tab:gpt_counselingeval} reports the evaluation of therapist skills in interactions with resistant clients. 

\paragraph{Text-based Versus Vision-augmented}
As can be seen, text-based LLMs generally struggled to engage with resistant clients, particularly in collaborative interactions that demand heightened sensitivity to client emotions.
This can be seen in the significant drop in performance compared to non-resistant clients.
In contrast, vision-enhanced models showed greater resilience, maintaining higher scores even when interacting with resistant clients. These results highlight the importance of nonverbal cues in effectively managing challenging client interactions. 

\paragraph{Fine-Tuning and CoT on CBT Performance}
Compared to \textsc{LLaVA-v1.5-7B}, which is the backbone model of \textsc{Mirror-LLaVA}, the \textsc{Mirror-LLaVA} family models achieved significantly higher scores in CBT-specific skills. This demonstrates the effectiveness of the \textsc{Mirror} dataset in enhancing CBT skills and reinforces the notion that, despite being trained on vast amounts of pre-existing data, LLMs still require targeted fine-tuning to effectively internalize and apply CBT principles. 
Further performance gains were observed when CoT processes, such as \textit{planning} and \textit{emotional captioning}, resulting in responses that were more contextually appropriate and emotionally attuned to the client’s needs.

\paragraph{Analysis of Response Length}

Excessively long response generation has been a persistent issue for LLMs and is known to reduce user satisfaction \cite{length}. Our analysis of response length revealed that, with the exception of the fine-tuned CBT counseling models (i.e. \textsc{Camel-LLaMA3} and \textsc{Mirror-LLaVA} family models), most models generated responses exceeding 30 tokens, which can degrade the counseling effectiveness. To further investigate these results, we provide actual examples for each model in Appendices~\ref{sec:case_study_length} and~\ref{sec:case_study_comparison_models}, and conduct an error analysis in Appendix~\ref{sec:error_analysis}.

\input{tables/gpt_TA}

\subsection{Client Alliance Assessment}

Table~\ref{tab:gpt_TA} presents the client alliance assessment using \textsc{GPT-4o}, which evaluates how well each model supports goal completion, establishes rapport (Approach), and fosters emotional connection (Affective Bond).

While overall alliance scores improve with \textsc{Mirror}, we observe a modest decline in the "Goal" score for \textsc{Mirror-LLaVA} models compared to some baselines. 
We attribute this to the design of the \textsc{Mirror} dataset, which emphasizes emotional engagement and rapport-building in resistant counseling scenarios, rather than directive goal setting.
In real-world counseling, especially under resistance, it is often more effective to prioritize emotional engagement before directive goal-setting. 
This trade-off is reflected in the substantial gains in Approach and Affective Bond scores, which more directly capture the model’s capacity for empathy and responsiveness.
Notably, \textsc{Mirror-LLaVA}$_{P + EC}$ achieves the highest scores in these affective dimensions, demonstrating the strength of step-by-step reasoning in managing resistance.

\input{figures/human_wr_ta}


\subsection{Domain Expert Assessment}
\label{sec:expert_assessment}

To further validate previous client alliance results, we conducted pairwise comparisons between \textsc{Mirror-LLaVA}$_{P+EC}$, \textsc{LLaMA-3-8B}, and \textsc{Camel-LLaMA3} using 200 randomly selected cases from the test set, balanced across three resistance categories: emotional, cognitive, behavioral. 
Specifically, two domain experts evaluated the models and selected the better model in each comparison (Appendix~\ref{sec:numerical_result}).

Moreover, we focused on comparing our method against the strongest baselines in CBT counseling—\textsc{LLaMA-3-8B} and \textsc{Camel-LLaMA3}, which are ranked highest in CBT-specific skill. 
Figure~\ref{fig:human_wr_ta} shows the average win rate across all pairwise comparisons. 
As depicted, the win rate confirmed that \textsc{Mirror-LLaVA}$_{P+EC}$ consistently outperformed its counterparts across all three dimensions of the therapeutic alliance. 
This result confirms that our model is not only favored in automatic evaluations but also by actual counseling experts.

In particular, while GPT-based evaluation showed limited gains in the "Goal" dimension, domain experts more frequently selected \textsc{Mirror-LLaVA}$_{P+EC}$ as superior in goal-related dialogue segments. 
Experts noted that goal pursuit was achieved more implicitly through sustained rapport and motivational alignment, rather than through direct or premature intervention. 
This reinforces our claim that emotional connection should precede goal-setting in resistant counseling contexts.

It is interesting to note that \textsc{GPT-4o}’s evaluation in Table~\ref{tab:gpt_TA} ranked \textsc{LLaMA-3-8B} higher than other LLMS, whereas domain experts preferred  \textsc{Camel-LLaMA3} significantly more in pairwise comparisons.
This discrepancy is likely due to human preference for responses of a more natural length, rather than those that are excessively long.


\input{figures/real_therapy}

\subsection{Application in Real-World Counseling Demonstrations}
\label{sec:application}

While our primary experiments relied on synthetic multimodal dialogues, we also sought to examine whether \textsc{Mirror}$_{P+EC}$ generalizes to more realistic settings. 
Since no CBT-based counseling video datasets are publicly available, we instead utilized motivational interviewing (MI) demonstration videos.
Specifically, we drew on the AnnoMI dataset \cite{9746035}, which contains 133 counseling session videos with aligned client transcriptions. 
For our demonstration, we focused on a subset of sessions in which clients displayed clear signs of resistance, as this aligns with the central aim of our study rather than providing a full evaluation of the entire dataset.

As shown in Figure \ref{fig:real_therapy}, clients exhibit various forms of resistance, including reluctance to seek help (Client A), claiming impunity (Client B), minimizing concerns (Client C), and externalizing blame (Client D)\footnote{These are well-documented resistance patterns in psychotherapy \cite{Miller2002-dm}.}. The case analysis demonstrates that \textsc{Mirror}$_{P+EC}$ effectively identifies the client’s emotional state through captioning and responds with emotional validation\footnote{This term refers to accepting a client's emotions without judgment, helping to sense of being understood while encouraging to manage their emotions.} and open-ended questions, common therapeutic techniques for managing resistance \cite{Miller2002-dm}. For example, when Client D externalizes blame onto family, the therapist acknowledges their feelings of isolation while gently redirecting the conversation toward exploring ways to cope with pressure. Further case studies can be found in Appendix \ref{sec:case_study_realworld}.

\section{Conclusion}
In this paper, we explore the use of multimodal cognitive reframing therapy for managing client resistance. 
Given the challenges faced by LLMs in addressing resistant clients and the potential advantages of VLMs, we aim to enhance AI therapists' ability to manage resistance by incorporating nonverbal cues, particularly facial expressions, to detect and understand client resistance. 
To address this challenge while mitigating privacy concerns associated with real-person data, we developed \textsc{Mirror}, a novel synthetic dataset for multimodal cognitive reframing therapy. 
Additionally, we have evaluated the AI therapist’s performance in two key areas: therapeutic skills and alliance-building, as well as adaptability to real-world counseling scenarios. 
Our results demonstrate significant improvements in both areas when trained with \textsc{Mirror}, underscoring its potential for real-world therapeutic applications. 
These improvements contribute to the development of AI therapists that are more empathetic and capable of fostering stronger therapeutic relationships.

\section*{Limitations}

\paragraph{Biases in Image Generation}
We used LLMs to generate image prompts, which were then rendered into facial images using PhotoMaker, grounded in the CelebA dataset.
However, both the language models and CelebA carry cultural and demographic biases that may have influenced the resulting images \cite{alkhamissi-etal-2024-investigating, naous-etal-2024-beer}.
For example, when prompted with “\textit{[Smiling] Hi,}” the LLM may describe a smile as “eyes curved like a crescent moon,” reflecting an East Asian view, whereas other cultures emphasize teeth or dimples \cite{8580373}.

Moreover, because facial identities were randomly sampled from CelebA and preserved using ArcFace embeddings, the demographic distribution of our dataset largely reflects that of CelebA, which is skewed toward faces of Western individuals \cite{bahng2020exploring}.
This likely reinforces existing imbalances inherent in the source dataset. 
As a result, \textsc{Mirror} does not ensure racial diversity and may underrepresent certain racial or ethnic groups, an important concern given that race and ethnicity affect both mental health outcomes and model performance \cite{meyer2013influence, doi:10.1177/1354067X231156600}.  
In addition, the strict image filtering step may inadvertently favor facial identities or expressions that are easier to preserve across variations \cite{pena2021facial, yucer2020exploring}, introducing subtle demographic or aesthetic biases into the final dataset.

Future work may explore multimodal counseling datasets with more balanced demographics to better capture cross-cultural variation in emotion and resistance.

\paragraph{Authenticity and Diversity of Nonverbal Cues}
Because the final facial images in \textsc{Mirror} were synthesized using an AI-based generation model, they may not fully capture the authenticity of real-world expressions. 
Subtle nuances such as micro-expressions, muscle tension, or natural asymmetries can be lost or inaccurately rendered, limiting the reliability of these cues for therapeutic interpretation. 
Moreover, resistance in counseling is expressed not only through facial expressions but also through body posture, voice tone, speech timing, and other multimodal signals. 
Since our study focused primarily on facial expressions and utterances, this scope restricts the representational diversity of resistant behaviors. 
Although the nonverbal cues in \textsc{Mirror} were grounded in behavioral traits commonly associated with resistance and informed by prior psychological literature \cite{chung2012impact}, further empirical validation is needed to establish their clinical accuracy and consistency. 
Future work could build on real-world video datasets (e.g., AnnoMI) to better align client utterances with authentic human behavior, incorporate audio-based nonverbal cues, and ultimately construct CBT-specific multimodal resources for more comprehensive modeling of resistance.

\paragraph{Scope and Session Length}
In contrast to typical counseling sessions, which last about an hour and extend over multiple interactions, our dataset consists of relatively short, single-session dialogues. 
This limitation makes it difficult to capture longer-term therapeutic processes such as sustained reframing or the gradual resolution of cognitive distortions. 
Moreover, all sessions are conducted in English, leaving open the question of how well models trained on \textsc{Mirror} would generalize to multilingual counseling settings. 
Future work could extend the dataset to longer, multi-session interactions and incorporate multilingual support to enable broader and more realistic applications.

\paragraph{Conversational Structure and Termination}
Our framework does not impose strict turn-level constraints or predefined termination points within the dialogue. 
While we incorporate a counseling strategy, \textit{planning}, to maintain goal orientation, the absence of explicit session boundaries may result in prolonged interactions without meaningful therapeutic progress. 
For example, if a simulated client remains in a negative emotional state, the AI therapist may continue offering supportive statements rather than facilitating cognitive change. 
This limitation highlights the importance of incorporating clearer session structures or exit strategies in future designs to better align with therapeutic goals.

\paragraph{Limited Expert Involvement in \textsc{Mirror}}
Although \textsc{Mirror} incorporates eight filtering steps—covering image-text similarity, identity and gender preservation, therapeutic alliance, safety, and other quality checks—these were largely adapted from prior work and do not fully substitute for direct expert supervision. 
In particular, while our WAI-based filtering step benefited from evidence showing moderate agreement between GPT-based and human expert evaluations (ICC $\approx$ 0.66--0.72) \cite{li-etal-2024-understanding-therapeutic}, the dataset creation process did not include human experts in the loop, primarily due to budgetary and privacy constraints. 
This absence represents a limitation, as human-in-the-loop supervision remains critical for ensuring clinical fidelity in psychotherapy research. 
Future work should incorporate expert-driven criteria derived from the psychotherapy literature, for example, filtering dialogues based on demonstrated empathy, which is a key component of CBT effectiveness. 
In addition, recent work such as \citet{liu2025eeyore} highlights how synthetic clients can be better aligned with real-world behaviors, offering a valuable direction for enhancing dataset realism.

\paragraph{Model Selection and Generalization} 
Although we trained the \textsc{LLaVA-v1.5-7B} model with two different CoT options and demonstrated its strong performance in handling client resistance and CBT counseling, our evaluation was based on a single backbone model. 
This could be a limitation, as there may be other VLMs that could perform better or differently, depending on their architecture or training. 
The reliance on a single model limits the generalizability of our findings. 
Future work may extend this line of research by not only comparing multiple VLM architectures, but also benchmarking against licensed human therapists. 
Such comparisons, for example using WAI scores, could highlight where AI systems align with or fall short of human therapists in managing resistance.

\paragraph{GPT-Based Evaluation}
Our primary evaluation relied on counseling sessions between a GPT-based client and AI therapists trained on the \textsc{Mirror} dataset. 
These conversations were evaluated using GPT-4o within two frameworks: \textsc{CounselingEval} (for general and CBT-specific skills) and a WAI-based framework for therapeutic alliance. 
While this setup enabled scalable and systematic analysis, it also introduces several limitations. 
GPT-4o showed a clear length bias in the “Strategy” dimension (Appendix \ref{sec:gpt_analysis}), and in the “Goal” dimension it produced similar scores across models, even though licensed therapists consistently preferred \textsc{Mirror}-based responses. 
This suggests that GPT-4o may under-recognize relational qualities such as empathy, underscoring that GPT-based evaluation cannot replace clinical standards. 
To address this, we adopted a layered approach that combines GPT-based breadth with human expert judgments on a subset of outputs. 
Although prior work has shown moderate agreement between GPT and expert ratings \cite{li-etal-2024-understanding-therapeutic}, such scores should be viewed as supportive rather than definitive.
Finally, due to cost and privacy constraints, we could not conduct large-scale human evaluations, but we supplemented our analysis with qualitative demonstrations on real counseling videos from the AnnoMI dataset (\S~\ref{sec:application}). 
Future work should integrate clinical expertise more directly to align evaluations with therapeutic outcomes.


\section*{Ethical Statement}

\paragraph{Privacy Considerations for Images} Ensuring privacy and ethical integrity is a fundamental priority in our dataset construction. We utilize the CelebA dataset \cite{liu2015faceattributes}, which is distributed under the MMLAB license. This license strictly prohibits commercial use and redistribution of the dataset. In compliance with these terms and to respect the rights of the individuals depicted in the images, we do not share the raw images directly. Instead, we provide image links and code that enables researchers to process the dataset independently, ensuring that the dataset's usage remains within ethical and legal boundaries.

\paragraph{Privacy Considerations for Dialogue}
The dialogue seeds for this dataset were sourced from the \textsc{Cactus} dataset \cite{lee-etal-2024-cactus}, with \textsc{PatternReframe} \cite{patternreframe} serving as its seed dataset. This dataset does not contain actual medical records but was collected through crowdsourcing, where each participant was assigned a persona and instructed to role-play. Additionally, during the dataset generation process, no utterances were derived from real individuals' personas; instead, all dialogues were fully synthesized. This approach further mitigates privacy concerns by ensuring that no personal data is incorporated into the dataset.

\paragraph{Safety Considerations} While AI has the potential to provide support, it may also have unintended negative effects on individuals with mental health challenges \cite{luxton2014recommendations}. Although our model has demonstrated some degree of effectiveness, our primary objective was to explore whether AI can effectively engage with patients who exhibit resistance to therapy. Therefore, we believe that AI should be used under the supervision of a professional rather than serving as a standalone tool in counseling sessions, particularly for individuals with severe psychological conditions beyond its intended scope. Additionally, to ensure the safety and appropriateness of the dataset, we implemented NSFW filtering and incorporated Canary to identify and remove conversations that may require human intervention.

\paragraph{Bias Considerations}
Although we utilize randomly selected images and a dialogue seed dataset that incorporates diversity in age, gender, and occupation, there remains a possibility of bias in our dataset. This is primarily due to our reliance on LLMs, which are predominantly trained on Western-centric datasets. 
In particular, during the screenplay generation process, gestures and nonverbal cues may vary across cultures. 
Since these were generated using \textsc{GPT-4o-mini}, certain gestures may not align with cultural norms in specific regions. Therefore, to ensure cultural appropriateness, retraining and adaptation would be necessary before deploying the model in a specific country.

\section*{Acknowledgements}
This work was supported by the  IITP(Institute of Information \& Coummunications Technology Planning \& Evaluation)-ITRC(Information Technology Research Center) grant funded by the Korea government(Ministry of Science and ICT)(IITP-2025-RS-2024-00437866, 47.5\%) and Smart HealthCare Program funded by the Korean National Police Agency(KNPA) (No. RS-2022-PT000186, 47.5\%), and Institute of Information \& Communications Technology Planning \& Evaluation (IITP) grant funded by the Korea government (MSIT) (No. RS-2019-II191906, Artificial Intelligence Graduate School Program (POSTECH), 5\%).

\bibliography{custom}

\clearpage

\appendix

\section{License}
\label{sec:license}
\textsc{Mirror} is constructed using the CelebA  \cite{liu2015faceattributes} and the \textsc{Cactus} datasets \cite{lee-etal-2024-cactus}. 
CelebA is released under the MMLAB license, which restricts redistribution, while \textsc{Cactus} is licensed under the GPL-2.0 license, permitting non-commercial scientific use. 
In adherence to these licensing terms, we do not directly include images from these datasets in \textsc{Mirror}. 
Instead, we provide links to the original sources. 
Consequently, \textsc{Mirror} is distributed under the GPL-2.0 license, ensuring compliance with the licensing conditions of the datasets used.

\section{Related Work}
\label{sec:related_work}
Research on AI-assisted cognitive reframing therapy has largely focused on text-based approaches with LLMs.
Early studies explored sentence rewriting to address cognitive distortions \cite{ziems-etal-2022-inducing, maddela-etal-2023-training, sharma-etal-2023-cognitive, goel2024socratic}, drawing on evidence that low-intensity CBT interventions can be effective in self-help formats \cite{Williams_2001, Shafran2021-cy}.
Subsequent work shifted toward conversational settings, evolving from simple query-response interactions \cite{na-2024-cbt, liu2023chatcounselor} to structured, multi-turn frameworks.
For example, \citet{xiao-etal-2024-healme} proposed a three-stage counseling process to ensure that AI functions as a facilitator rather than a direct corrector. 
Other studies have emphasized improving the realism of cognitive reframing datasets \cite{lee-etal-2024-cactus} and enhancing AI therapists' professional counseling competence \cite{zhang-etal-2024-cpsycoun}.

Most recently, there has been a growing interest in incorporating nonverbal cues into AI-driven psychotherapy. 
Building on \citet{xiao-etal-2024-healme}, \citet{kim-etal-2025-multimodal} investigated multimodal cognitive reframing, showing that VLMs can generate more empathic responses.
Similarly, \citet{10.1145/3581783.3612346} introduced a multimodal empathy dataset in counseling, underscoring the role of clients’ nonverbal expressions. 
In parallel, \citet{kebe-etal-2025-llamadrs} demonstrated that even text-only LLMs can approach human reliability in depression assessment, while highlighting the difficulty of evaluating symptoms that depend on nonverbal cues—further motivating multimodal approaches.

Other recent efforts include the release of specialized resources for psychotherapy, such as \textsc{CBT-Bench} \cite{zhang-etal-2025-cbt}, a benchmark for evaluating AI therapists’ CBT capabilities, and \textsc{MIDAS} \cite{gunal-etal-2025-examining}, a Spanish motivational interviewing (MI) video dataset that expands multilingual multimodal research.

Our work advances multimodal cognitive reframing by focusing on managing client resistance, strengthening the therapeutic alliance, and improving AI-assisted psychotherapy.

\section{Training Details}
\label{sec:training}
The \textsc{LLaVA-v1.5-7B} model was fine-tuned on the \textsc{Mirror} dataset using LoRA \cite{hu2022lora} for 5 epochs. 
We used the official \textsc{LLaVA-v1.5-7B} model from Hugging Face\footnote{\url{https://huggingface.co/liuhaotian/llava-v1.5-7b}} and followed the default hyperparameters\footnote{\url{https://github.com/haotian-liu/LLaVA/tree/main}}, which include a learning rate of 2e-5, an AdamW optimizer without weight decay, and a cosine learning rate schedule with a 3\% warmup ratio.
Training was done on four A100-80GB GPUs with a batch size of 32 per GPU.

\input{figures/correlation_w_length}

\section{Impact of Response Length on GPT Evaluation}
\label{sec:gpt_analysis}

We examined the correlation between the AI therapist’s response length and its performance in GPT-based evaluation. Across all models, we analyzed how response length affects evaluation metrics and further aggregated the results by modality. As shown in Figure \ref{fig:general_performance_analysis} and Figure \ref{fig:cbt_performance_analysis}, there is a noticeable relationship between response length and performance in both general counseling skills and CBT techniques.
Notably, the strongest correlation was observed in the Strategy category, with a correlation of 0.6, suggesting that untrained text-based LLMs tend to receive higher evaluations from the GPT evaluator when generating longer responses. 
This is likely because lengthier responses incorporate multiple questions or strategies within a single reply, which the evaluator interprets as higher-quality output. In contrast, for VLMs, response length showed no significant correlation with performance in general counseling skills. 

However, within CBT techniques, particularly in Focus and Guided Discovery, shorter responses generally resulted in higher scores. 
This trend is likely influenced by the \textsc{LLaVA-v1.5-7B} model, which tends to generate unnaturally long responses and has lower scores. Compared to the \textsc{LLaVA-v1.5-7B} model, the \textsc{Mirror-LLaVA} family produced shorter responses and achieved better scores, suggesting a correlation between shorter responses and higher performance.

\section{Domain Expert Assessment Details}
\label{sec:numerical_result}
\subsection{Numerical Details}
Table~\ref{tab:human_wr1}, ~\ref{tab:human_wr2}, and ~\ref{tab:human_wr3} show the winning rates for each metric: Goal, Approach, and Affective Bond.
\subsection{Domain Expert Recruitment}
For the domain expert evaluation, we hired two evaluators through the Upwork platform\footnote{www.upwork.com} who hold a counseling license or have a graduate degree in a related field. They were informed that all personal information would remain anonymous and that responses would be used solely for research purposes. We paid \$0.05 per data entry for pairwise comparison, which they accepted before proceeding with the task.

\input{tables/numerical_results}

\section{Error Analysis}
\label{sec:error_analysis}

To gain deeper insights into the effectiveness and limitations of our proposed method, we conducted an error analysis on the \textsc{Mirror-LLaVA}$_{P+EC}$ model, and focused on cases with therapist skill and client alliance scores of less than 3. 

\subsection{Failure Cases in General Counseling Skills}

\input{figures/error_general_counseling}

Figure \ref{fig:error_analysis_a} illustrates failure cases in general counseling skills. The first case (Figure \ref{fig:general_counseling_a}) is due to a hallucination from the VLM. Although the client did not mention that a colleague was a pedophile, the VLM therapist incorrectly introduced this idea, which made the client uncomfortable. This misstep resulted in low understanding and interpersonal skills.

The second case (Figure \ref{fig:general_counseling_b}) involves a client who expressed deep-seated fear and emotional reluctance, stating,\textit{ ``I'm always worried about saying or doing the wrong thing.''} Rather than further exploring the client's underlying concerns, the model prematurely attempted to fix the problem before building intimacy. This response failed to align with the client's emotional state, leading to disengagement, which highlights the need for more specific \textit{planning} in CBT counseling.

\subsection{Failure Cases in CBT-Specific Skills}

\input{figures/error_cbt_specific}

Figure \ref{fig:error_analysis_b} illustrates failure cases in CBT-specific skills. In the first case, shown in Figure \ref{fig:cbt_a}, confusion between the therapist's and the client's roles occurred. In this case, the therapist's utterance shifted to client's utterance in one turn. Although this happened in only five cases, it resulted in a drop in the focus score.

The second case (Figure \ref{fig:cbt_b}) arises when the therapist loses their purpose and simply sympathizes with the client’s cognitive distortions. Instead of actively challenging the client's distorted thought patterns, the model engaged in emotion-focused inquiry, asking about specific experiences related to the client's feelings. While this approach may encourage emotional processing, it falls short in fostering cognitive reframing. A more effective intervention would involve helping the client examine the reasons behind their beliefs and exploring alternative perspectives.

\subsection{Failure Cases in Client Alliance}

\input{figures/error_TA}

In analyzing cases where the client alliance score was below 3, we identified a key issue that hinders effective therapeutic engagement (Figure \ref{fig:error_ta}). In this example, the client expresses a strong sense of reluctance and feeling stuck in a negative mindset, signaling deep-seated emotional resistance. 
However, rather than exploring these emotions further, the therapist prematurely shifts the focus toward finding positive experiences. 
While encouraging positive reframing is valuable, doing so too soon leads to a mismatch in attunement, making the client feel unheard or dismissed.

\section{Case Study}
\label{sec:case_study}
We conducted a comparative analysis of AI therapist counseling sessions to examine how different models respond to and handle client resistance.

\subsection{Response Length Analysis Across Models}
\label{sec:case_study_length}
\input{figures/case_study_length}

We compare the response lengths of different models when a client expresses the distorted thought, ``\textit{bad things will happen}''. Figure \ref{fig:case_length} presents the actual responses from five AI therapist models to a virtual client's statement exhibiting cognitive distortion.
The models that were not fine-tuned with CBT datasets, including \textsc{LLaMA-3-8B}, \textsc{GPT-3.5-Turbo}, and \textsc{LLaVA-v1.5-7B}, tended to generate excessively long responses, which negatively impacted the effectiveness and naturalness of the dialogue. In contrast, \textsc{Camel-LLaMA3} and \textsc{Mirror-LLaVA}$_{P+EC}$, which were fine-tuned for CBT counseling, produced responses of more appropriate lengths, showing better alignment with client needs and making the interactions easier for clients to understand.

\subsection{Comparison of Resistance Management in Virtual Counseling}
\label{sec:case_study_comparison_models}
\input{figures/case_study}

We conducted a comparative analysis of two counseling-optimized models, \textsc{Mirror-LLaVA}$_{P+EC}$ and \textsc{Camel-LLaMA3}, to evaluate their approaches to handling client resistance in virtual counseling sessions. Both models share the same goal of CBT and incorporate \textit{planning} during inference. However, \textsc{Camel-LLaMA3} is trained on a general CBT dataset and does not specifically address client resistance. In the \textsc{Camel-LLaMA3} session (Figure \ref{fig:case_camel}), when the client exhibited resistance, the model primarily relied on emotional validation and exploration. While these techniques offer comfort to client, they do not challenge deeper, malformed beliefs. 
As a result, while this model focused on emotional exploration with surface-level validation of the client’s negative emotions, its reframing process lacked progression.

In contrast, \textsc{Mirror-LLaVA}$_{P+EC}$ (Figure \ref{fig:case_mirror}) employed a more nuanced approach, integrating emotional validation, positive reinforcement, and cognitive reframing. Despite the client’s resistance, our model attempted to delicately reframe the client’s thoughts, using a collaborative approach with statements such as, ``\textit{We can work on identifying those thoughts and reframing them into something more empowering together.}'' Furthermore, by asking questions like “\textit{How does that sound?}” the model encouraged client engagement and showed respect for the client’s perspective. The model also effectively used positive reinforcement to encourage clients who were hesitant to take action by offering supportive statements like, “\textit{That’s a brave and important step}”. These findings underscore the importance of specialized training datasets for client resistance, such as \textsc{Mirror}, in effectively managing resistance and fostering therapeutic growth.

\subsection{Real-World Counseling Demonstrations}
\label{sec:case_study_realworld}
Here, we provide additional explanation for the cases in Figure~\ref{fig:real_therapy}. For Client A, the repeated use of `\textit{I don't know}' illustrates reluctance to seek help, indicating emotional uncertainty and a lack of motivation to engage in the process. However, our model effectively addresses this resistance by validating the client's feelings and gently encouraging exploration of their concerns, thereby guiding the client toward self-awareness and understanding.

For Client B, the client initially exhibits a sense of impunity regarding their drinking habits, reflecting the distorted thought that `\textit{everyone drinks like me}' which can make cognitive reframing challenging. However, our model successfully recognizes the client's uncertainty and potential for change and start to addresses the resistance by gently encouraging further exploration of their thoughts.

Lastly, for Client C, the client minimizes the concerns raised by others, expressing surprise at the intervention. Rather than confronting the client directly, the therapist takes a more empathetic approach by first acknowledging the client's feelings. This helps to build rapport and create a safe space, encouraging the client to open up for deeper, more effective counseling in future sessions.

\section{Effect of Stage Direction}
\label{sec:stage_direction}

\input{figures/data_case}
Stage directions, commonly used in theater to guide actors in terms of gaze, posture, and vocal tone, are applied in our approach so as to synthesize more realistic images. To assess the impact of incorporating these stage directions, we compared the results with and without facial image synthesis (\S\ref{sec:step3}). Figure \ref{fig:data_case} presents four examples that illustrate this comparison. By integrating cues such as gaze direction and arm positioning, the generated client images align more naturally with the intended speech, thereby enhancing both the realism and contextual relevance of the dataset.

\onecolumn

\section{Prompts for \mirror \textsc{Mirror}}
\label{sec:mirror_prompts}

\input{figures/dataset_prompt}

\clearpage 
\section{Prompts for Counseling Simulation}
\label{sec:counseling_prompts}

For baseline comparisons, we followed the official therapist simulation prompts and prompt structures from \citet{lee-etal-2024-cactus} for \textsc{Camel-LLaMA3}, \textsc{GPT-3.5-Turbo}, and \textsc{LLaMA-3-8B}, with the planning component in \textsc{Camel-LLaMA3} implemented as specified in their work.

\input{figures/counseling_prompts}

\clearpage

\section{Prompts for Evaluating AI Therapists}
\label{sec:evaluating_ai_therapist}

Therapist skills were assessed using the prompts provided in the official \textsc{CounselingEval} code\footnote{https://github.com/coding-groot/cactus} by \citet{lee-etal-2024-cactus}.

For client alliance assessment, we adopted the Working Alliance Inventory (WAI) questions as adapted by \citet{li-etal-2024-understanding-therapeutic}.
These consist of twelve items covering three dimensions: Goals, Approach, and Affective Bond, each rated on a 5-point scale. 
The detailed evaluation guidelines for each question follow those provided in \citet{li-etal-2024-understanding-therapeutic} and are not reproduced here for brevity.
Table~\ref{tab:wai_questions} summarizes the twelve questions grouped by dimension.

\begin{table}[h]
\centering
\renewcommand{\arraystretch}{1.2}
\begin{tabular}{c p{12cm}}
\hline
\textbf{Dimension} & \textbf{Question} \\
\hline
\multirow{4}{*}{Goals} 
 & Q1. There is mutual understanding about what participants are trying to accomplish in therapy. \\
 & Q2. The client and counselor are working on mutually agreed upon goals. \\
 & Q3. The client and counselor have the same ideas about what the client’s real problems are. \\
 & Q4. The client and counselor have established a good understanding of the changes that would be good for the client. \\
\hline
\multirow{4}{*}{Approach} 
 & Q5. There is agreement about the steps taken to help improve the client’s situation. \\
 & Q6. There is agreement about the usefulness of the current activity in therapy. \\
 & Q7. There is agreement on what is important for the client to work on. \\
 & Q8. The client believes that the way they are working with his/her problem is correct. \\
\hline
\multirow{4}{*}{Affective Bond} 
 & Q9. There is a mutual liking between the client and counselor. \\
 & Q10. The client feels confident in the counselor’s ability to help the client. \\
 & Q11. The client feels that the counselor appreciates him/her as a person. \\
 & Q12. There is mutual trust between the client and counselor. \\
\hline
\end{tabular}
\caption{Client alliance assessment questions grouped by dimension.}
\label{tab:wai_questions}
\end{table}

The following template illustrates the prompt format we used to evaluate each question:

\input{figures/eval_prompts}

\clearpage 
\section{A Full Example of \mirror \textsc{Mirror}}
\input{figures/mirror_full}

\clearpage 
\section{Instruction for Human Pairwise Comparison}

\begin{figure*}[h!] 
    \centering
    \includegraphics[width=\textwidth]{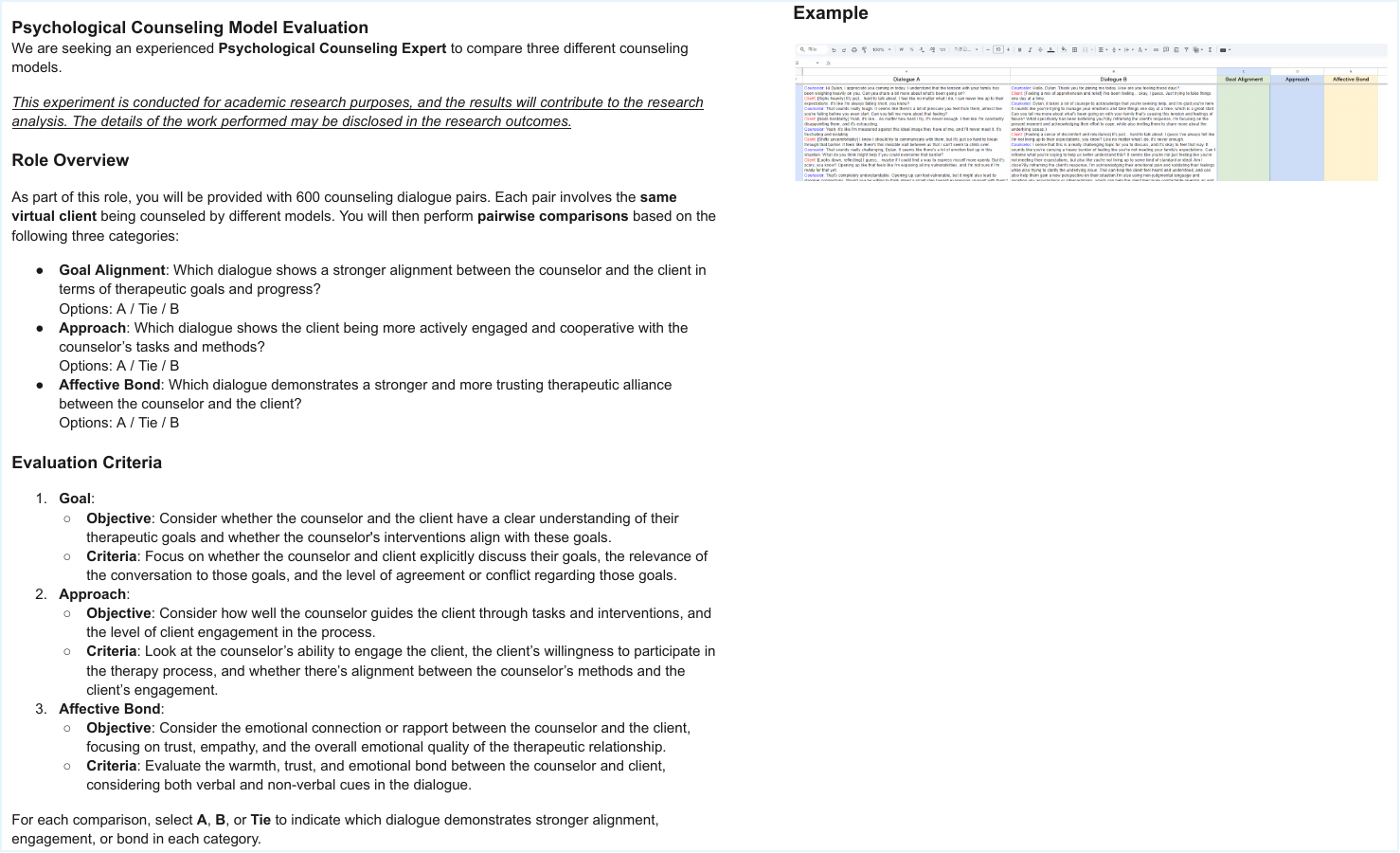} 
\end{figure*}

\end{document}

%% file: figures/problem.tex
\begin{figure}[t!]
\centering
  \includegraphics[width=\linewidth]{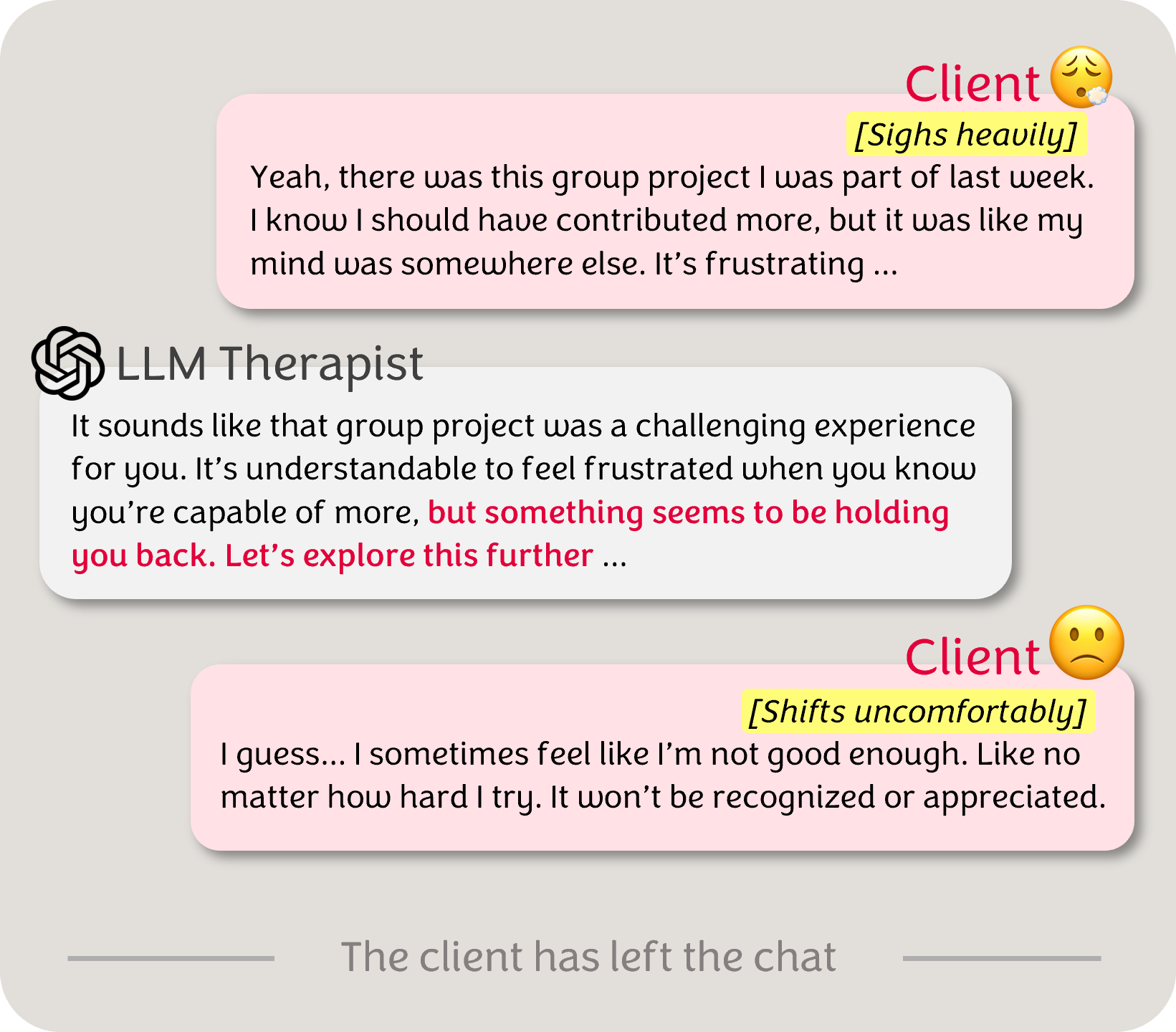}
  \caption{
  Text-based therapists have limitations in interpreting nonverbal cues, as they cannot perceive behaviors such as sighs or posture shifts, which can lead to premature problem-solving rather than addressing deeper emotions.
  }
  \label{fig:problem}
\end{figure}

%% file: figures/dataset.tex
\begin{figure*}[t!] 
    \centering
    \includegraphics[width=\textwidth]{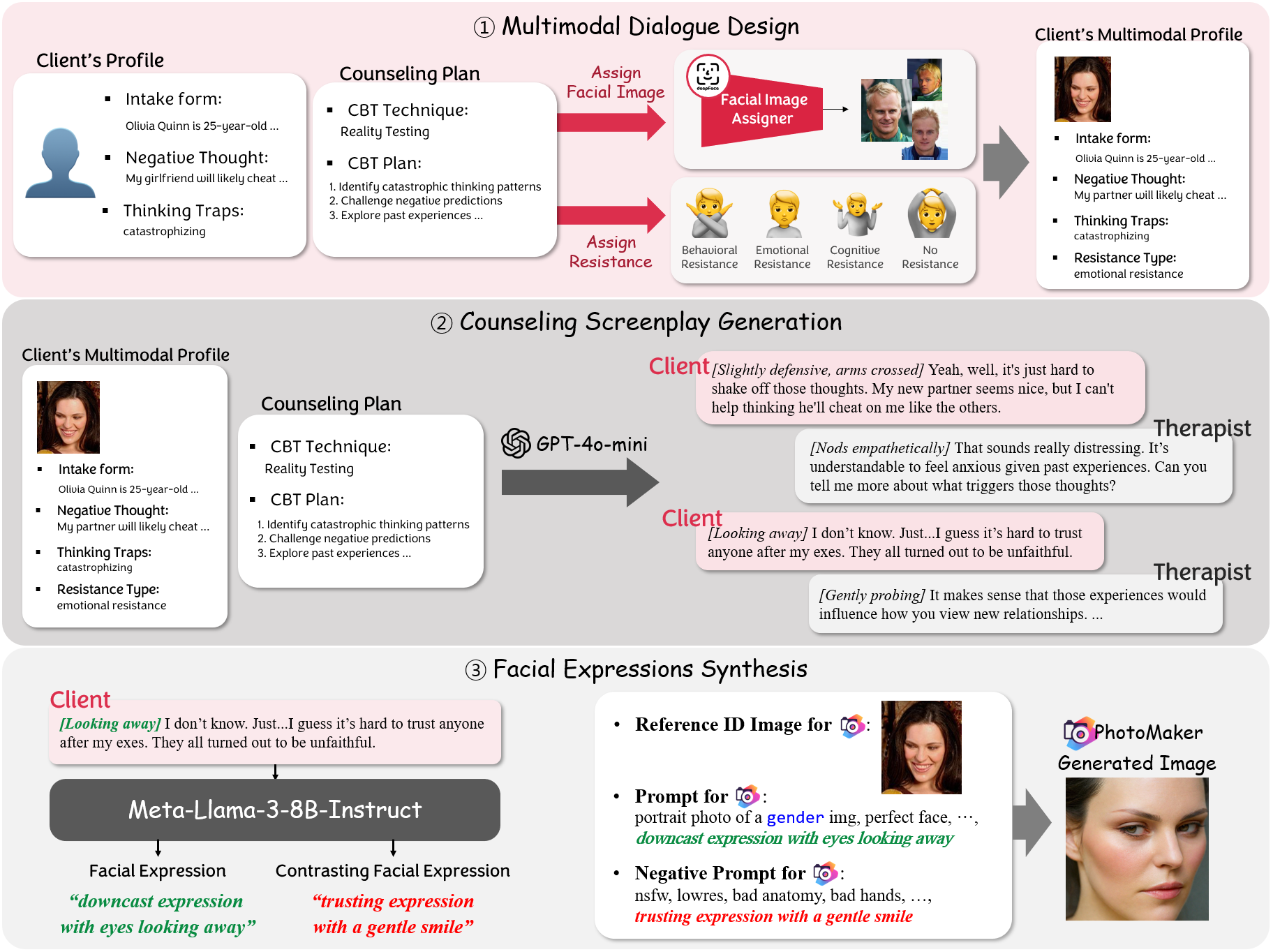} 
    \caption{Overview of the \textsc{Mirror} dataset construction. The pipeline consists of three main stages: Multimodal Dialogue Design (\S\ref{sec:step1}), Counseling Screenplay Generation (\S\ref{sec:step2}), and Facial Expression Synthesis (\S\ref{sec:step3}).} 
    \label{fig:dataset} 
\end{figure*}

%% file: tables/dataset_comparison.tex
\begin{table*}[t!]
    \centering
    \resizebox{\linewidth}{!}{
        \begin{tabular}{lcccccc}
            \toprule
            & \textbf{Modality} & \textbf{Language} & \textbf{\# of Dialogue} & \textbf{\# Avg. Turns}  & \textbf{\# Avg. Images} & \textbf{Turn-Image Alignment}\\
            \midrule
            Psych8k~\citep{liu2023chatcounselor}  & T & English & 8,187  & 1.00  & - &  -\\
            HealMe~\citep{xiao-etal-2024-healme} & T & English & 1,300 & 3.00 & - &  -\\
            \textsc{Cactus}~\citep{lee-etal-2024-cactus} & T & English & 31,577 & 16.6 & - & -\\
            CPsyCounD~\citep{zhang-etal-2024-cpsycoun} & T & Chinese & 3,134 & 8.7  & - & -\\
            M2CoSC~\citep{kim-etal-2025-multimodal} & T, V  & English & 429 & 4.00  & 1.00  & \xmark \\
            MEDIC~\citep{10.1145/3581783.3612346} & T, V, A  & Chinese  & 771 & 1.00 & 1,137 & \wmark\\
            \mirror\textsc{Mirror} & T, V & English  & 3,073 & 10.3 & 9.51 & \cmark\\
            \bottomrule
        \end{tabular}
    }
    \caption{
        Comparison of \textsc{Mirror} with other psychological counseling datasets. The \textbf{Modality} column indicates whether the dataset includes text (T), visual (V), or audio (A) data. \textbf{\# Avg. Images} refers to the average number of client images per dialogue. \textbf{Turn-Image Alignment} indicates whether the client images are dynamically aligned according to each dialogue turn.
        \wmark indicates that MEDIC, being single-turn data, cannot provide turn-level alignment of facial expressions as a session progresses.
    }
    \label{tab:dataset_comparison}
\end{table*}

%% file: figures/planning.tex
\begin{figure}[bt!] 
    \centering
    \includegraphics[width=\linewidth]{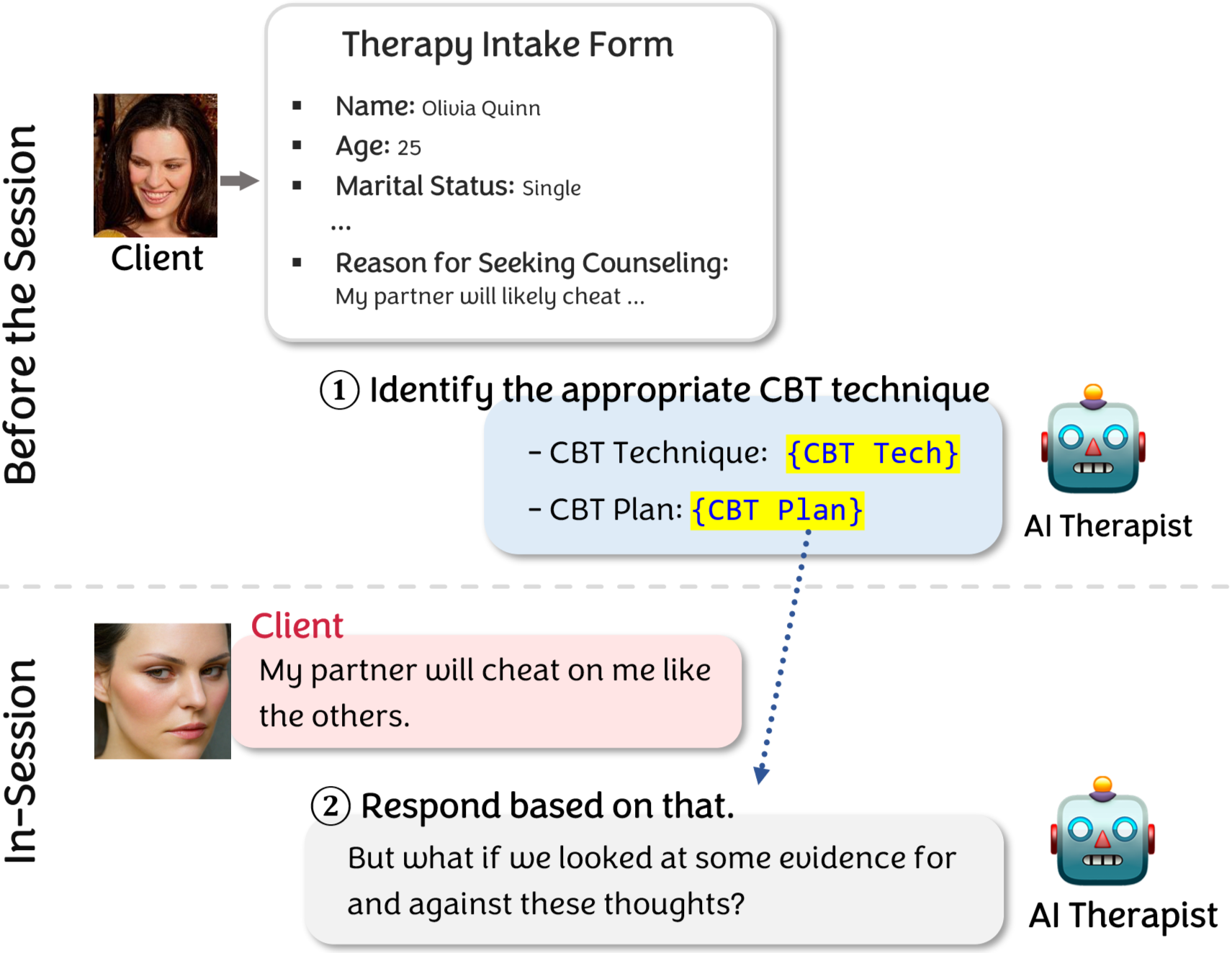} 
    \caption{
    The overview of the \textit{planning} process.
    } 
    \label{fig:planning} 
\end{figure}

%% file: figures/emotional_caption.tex
\begin{figure}[t!] 
    \centering
    \includegraphics[width=\linewidth]{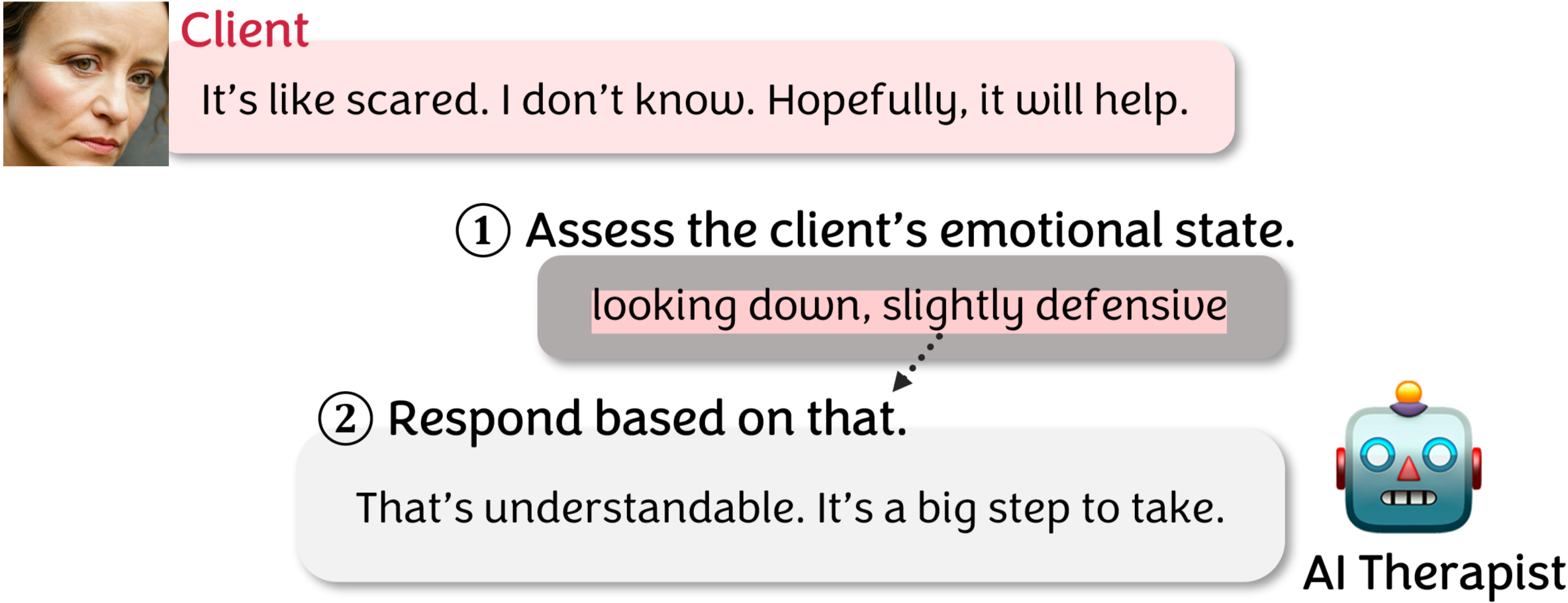} 
    \caption{
    Overview of \textit{emotional captioning}. The AI therapist infers the client’s emotional state from facial cues and uses it to generate an empathetic, aligned response.
    } 
    \label{fig:emotional_captioning} 
\end{figure}

%% file: tables/gpt_counselingeval.tex
\begin{table*}[t!]
  \centering
  \renewcommand{\arraystretch}{1.3}
  \resizebox{\linewidth}{!}{%
\begin{tabular}{>{\arraybackslash}m{3.85cm}|ccc|cc|cc}
    \Xhline{2.5\arrayrulewidth}
    \multirow{2}{*}{Model} & \multicolumn{3}{c|}{\textbf{General Counseling Skills ($\uparrow$)}} & \multicolumn{2}{c|}{\textbf{CBT-specific Skills ($\uparrow$)}} & \multicolumn{2}{c}{\textbf{Response Length}} \\ 
                            & Understanding  & Interpersonal Effectiveness & Collaboration & Guided Discovery & Focus & Avg. & Max \\ 
    \toprule
        \textsc{LLaMA-3-8B}  
            & 3.811$^*$\small{ $\vert$ {\color{blue}-0.073}}     
            & \textbf{4.114}\small{ $\vert$ {\color{blue}-0.012}}
            & 2.734$^*$\small{ $\vert$ {\color{blue}-0.311}}
            & 3.689$^*$\small{ $\vert$ {\color{blue}-0.096}} 
            & 3.692$^*$\small{ $\vert$ {\color{blue}-0.057}} 
            & 59.36 & 104.59 \\
        \textsc{Camel-LLaMA3}  
            & 3.794$^*$\small{ $\vert$ {\color{blue}-0.085}}
            & 4.003$^*$\small{ $\vert$ {\color{blue}-0.002}}      
            & 2.279$^*$\small{ $\vert$ {\color{blue}-0.198}}
            & 3.527$^*$\small{ $\vert$ {\color{blue}-0.127}} 
            & 3.563$^*$\small{ $\vert$ {\color{blue}-0.197}} 
            & 20.54 & 27.42\\
        \textsc{GPT-3.5-Turbo}       
            & 3.798$^*$\small{ $\vert$  {\color{blue}-0.172}}     
            & 4.049\small{ $\vert$ {\color{blue}-0.041}}       
            & 2.976$^*$\small{ $\vert$ {\color{blue}-0.194}} 
            & 3.462$^*$\small{ $\vert$ {\color{blue}-0.262}} 
            & 3.491$^*$\small{ $\vert$ {\color{blue}-0.238}} 
            & 36.19 & 57.28 \\ 
        \textsc{LLaVA-v1.5-7B}       
            & 3.622$^*$\small{ $\vert$ {\color{blue}-0.066}}  
            & 3.997$^*$\small{ $\vert$ {\color{red}+0.007}}     
            & 3.408$^*$\small{ $\vert$ {\color{red}+0.071}}
            & 2.494$^*$\small{ $\vert$ {\color{red}+0.057} }
            & 2.501$^*$\small{ $\vert$ {\color{blue}-0.012} }
            &  112.41 & 177.11\\
    \midrule
        \textsc{Mirror-LLaVA} 
            & 3.973$^*$\small{ $\vert$ {\color{blue}-0.017}}
            & 4.025\small{ $\vert$ {\color{blue}-0.040}}
            & 3.576$^*$\small{ $\vert$ {\color{blue}-0.089}}
            & 3.875$^*$\small{ $\vert$ {\color{red}-0.025}} 
            & 3.888$^*$\small{ $\vert$ {\color{red}-0.012}}
            & 27.68 & 32.14\\
        \textsc{Mirror-LLaVA}$_{P}$ 
            & 3.985\small{ $\vert$ {\color{blue}-0.015}} 
            & 4.098\small{ $\vert$ {\color{red}+0.063}} 
            & 3.722$^*$\small{ $\vert$ {\color{red}+0.117}}
            & 3.915$^*$\small{ $\vert$ {\color{blue}-0.040}}
            & 3.915$^*$\small{ $\vert$ {\color{red}+0.015}}
            &  27.00 & 32.02\\
        \cellcolor[HTML]{ffdbe0}\textsc{Mirror-LLaVA}$_{P + EC}$ 
            & \cellcolor[HTML]{ffdbe0}\textbf{4.000}\small{ $\vert$ {\color{red}+0.010}}
            & \cellcolor[HTML]{ffdbe0}4.055\small{ $\vert$ {\color{red}+0.010}}
            & \cellcolor[HTML]{ffdbe0}\textbf{3.913}\small{ $\vert$ {\color{blue}-0.082}}
            & \cellcolor[HTML]{ffdbe0}\textbf{3.977}\small{ $\vert$ {\color{red}+0.007}}
            & \cellcolor[HTML]{ffdbe0}\textbf{3.977}\small{ $\vert$ {\color{red}+0.037}}
            & \cellcolor[HTML]{ffdbe0}27.55 
            & \cellcolor[HTML]{ffdbe0}34.20\\ 
    \bottomrule
  \end{tabular}
   }
  \caption{\label{tab:gpt_counselingeval}
    Therapist skills assessment scores calculated by \textsc{GPT-4o} and response length. 
    Asterisk ($*$) indicates a significant difference compared to \colorbox[HTML]{ffdbe0}{\textsc{Mirror-LLaVA}\textnormal{$_{P + EC}$}} ($p$ < 0.05, paired t-test). 
     \textbf{Response Length} denotes the average and maximum number of tokens per turn. 
    Values after the vertical bar ($\vert$) indicate performance changes when interacting with resistant clients, relative to non-resistant clients; negative values denote a decline.
  }
  
\end{table*}

%% file: tables/gpt_TA.tex
\begin{table}[t!]
  \centering
  \renewcommand{\arraystretch}{1.2}
    
  \resizebox{\columnwidth}{!}{%
\begin{tabular}{>{\arraybackslash}m{3.8cm}|ccc}
    \Xhline{2.5\arrayrulewidth}
        \multirow{2}{*}{Model} & \multicolumn{3}{c}{\textbf{Client Alliance Skills ($\uparrow$)}}\\ 
    &  \textbf{Goal} & \textbf{Approach} & \textbf{Affective Bond}   \\ 
    \toprule
        \textsc{LLaMA-3-8B} 
            & 2.412$^*$\small{ $\vert$ {\textcolor{blue}{-0.023}}}
            & 3.309$^*$\small{ $\vert$ {\textcolor{blue}{-0.107}}}
            & 3.356$^*$\small{ $\vert$ {\textcolor{blue}{-0.138}}}\\ 
        \textsc{Camel-LLaMA3} 
            & 2.358$^*$\small{ $\vert$ {\textcolor{blue}{-0.009}}} 
            & 3.130$^*$\small{ $\vert$ {\textcolor{blue}{-0.072}} }
            & 3.149$^*$\small{ $\vert$ {\textcolor{blue}{-0.203}}}\\ 
        \textsc{GPT-3.5-Turbo} 
            & 2.472$^*$\small{ $\vert$ {\textcolor{blue}{-0.018}}}    
            & 3.272$^*$\small{ $\vert$ {\textcolor{blue}{-0.168}}} 
            & 3.297$^*$\small{ $\vert$ {\textcolor{blue}{-0.253}}} \\
        \textsc{LLaVA-v1.5-7B}
            & \textbf{2.589}\small{ $\vert$ {\textcolor{blue}{-0.048}}} 
            & 3.234$^*$\small{ $\vert$ {\textcolor{blue}{-0.181}}} 
            & 3.356$^*$\small{ $\vert$ {\textcolor{blue}{-0.163}}}\\ 
    \midrule
        \textsc{Mirror-LLaVA} 
            & 2.459$^*$\small{ $\vert$ {\textcolor{blue}{-0.033}}} 
            & 3.289$^*$\small{ $\vert$ {\textcolor{blue}{-0.060}}} 
            & 3.400$^*$\small{ $\vert$ {\textcolor{blue}{-0.092}}}\\ 
        \textsc{Mirror-LLaVA}$_{P}$
            & 2.525$^*$\small{ $\vert$ {\textcolor{red}{+0.033}}} 
            & 3.340\small{ $\vert$ {\textcolor{blue}{-0.005}}}
            & 3.448\small{ $\vert$ {\textcolor{blue}{-0.051}}}\\
        \cellcolor[HTML]{ffdbe0} \textsc{Mirror-LLaVA}$_{P + EC}$
            & \cellcolor[HTML]{ffdbe0} 2.567\small{ $\vert$ {\textcolor{red}{+0.035}}}
            & \cellcolor[HTML]{ffdbe0} \textbf{3.366}\small{ $\vert$ {\textcolor{blue}{-0.003}}}
            & \cellcolor[HTML]{ffdbe0}\textbf{3.480}\small{ $\vert$ {\textcolor{blue}{-0.024}}}\\

    \bottomrule
  \end{tabular}
  }
  \caption{
    Client alliance assessment results as evaluated by \textsc{GPT-4o}. 
  }
  \label{tab:gpt_TA}

\end{table}

%% file: figures/human_wr_ta.tex
\begin{figure}[t!]
\centering
  \includegraphics[width=\columnwidth]{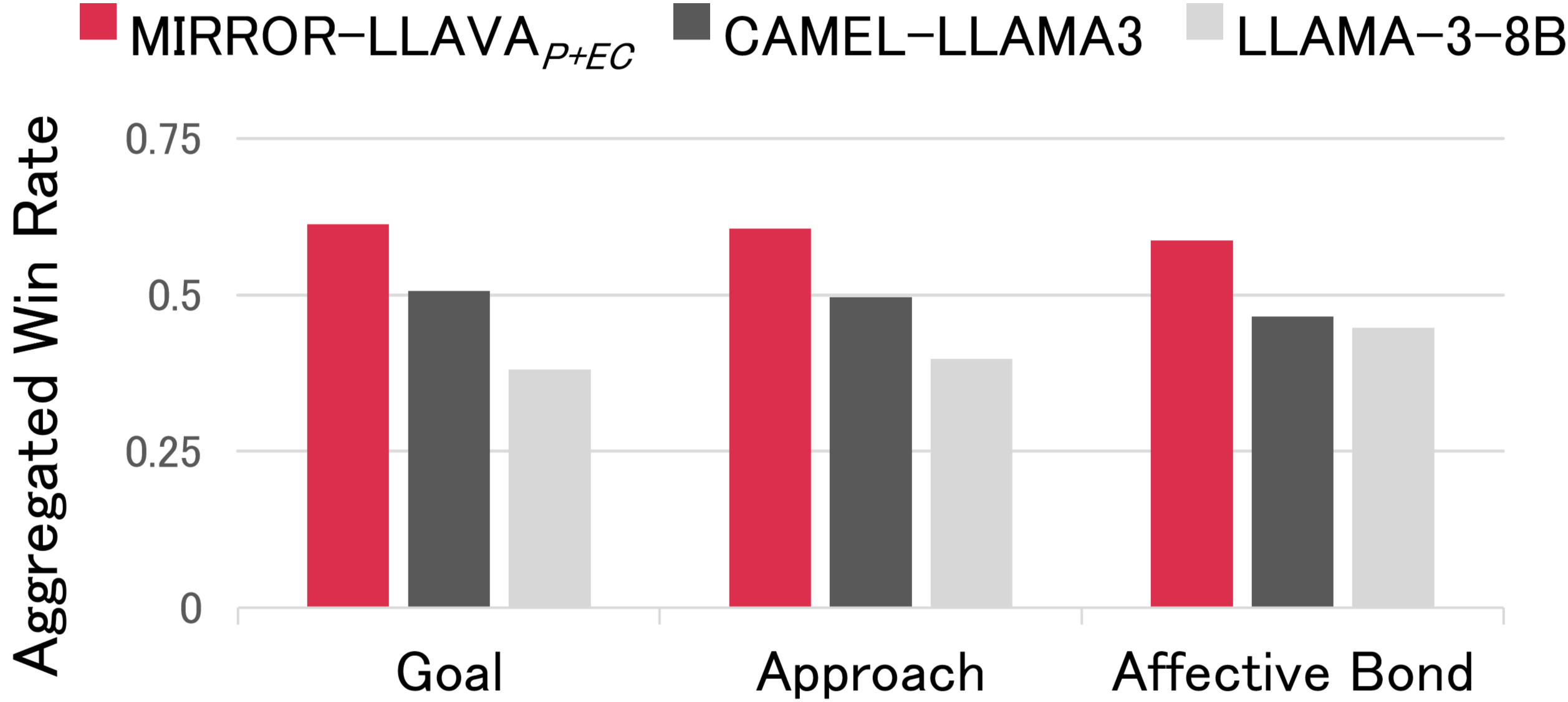}
  \caption{Pairwise comparison results among \textsc{Mirror-LLaVA}, \textsc{Camel-LLaMA3} and \textsc{LLaMA-3-8B}, on three evaluation criteria—Goal, Approach, and Affective Bond—rated by two psychotherapists.}
  \label{fig:human_wr_ta}
\end{figure}

%% file: figures/real_therapy.tex
\begin{figure}[t!]
    \centering
    \begin{subfigure}[t]{\linewidth}
        \centering
        \includegraphics[width=\linewidth]{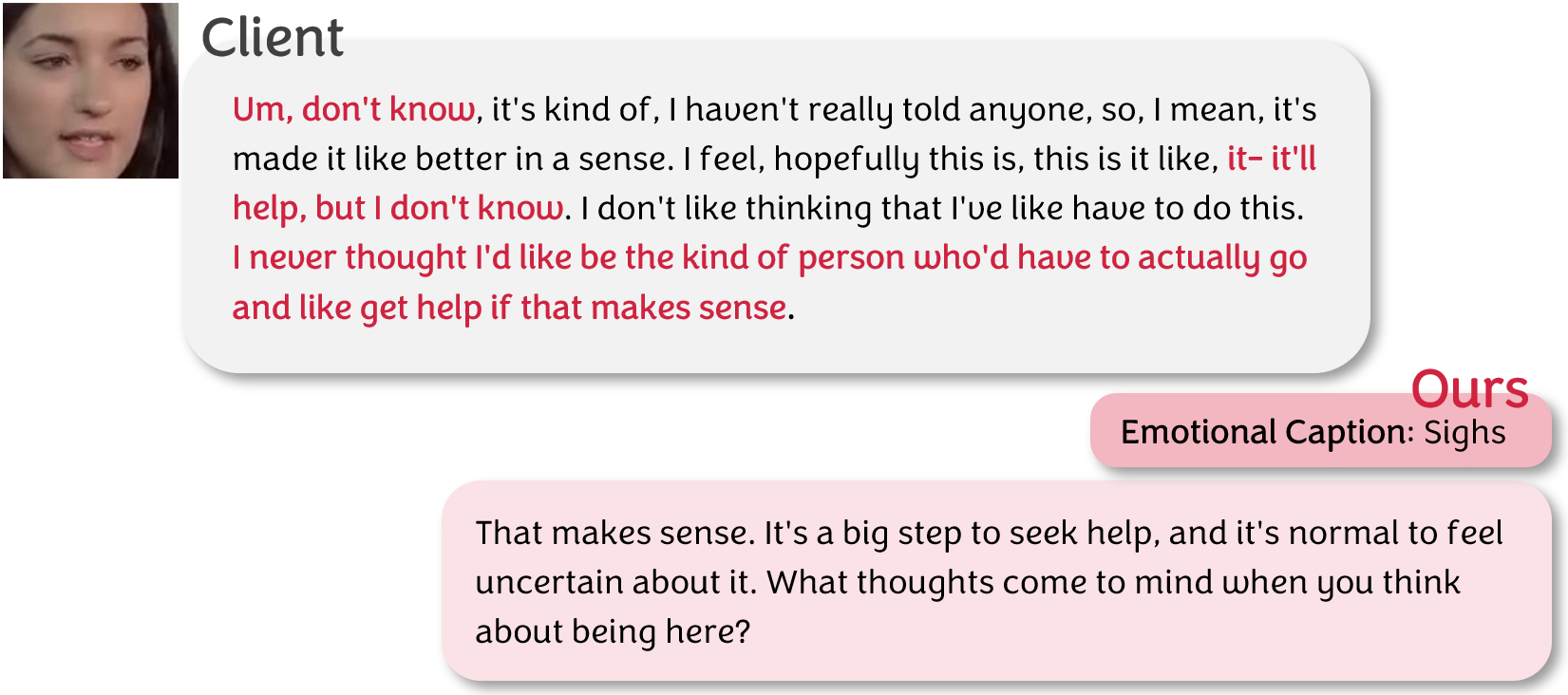} 
        \caption{Client A case: Reluctance to seek help}
        \label{fig:client_a}
    \end{subfigure}
    
    \vspace{2mm}
    
    \begin{subfigure}[t]{\linewidth}
        \centering
        \includegraphics[width=\linewidth]{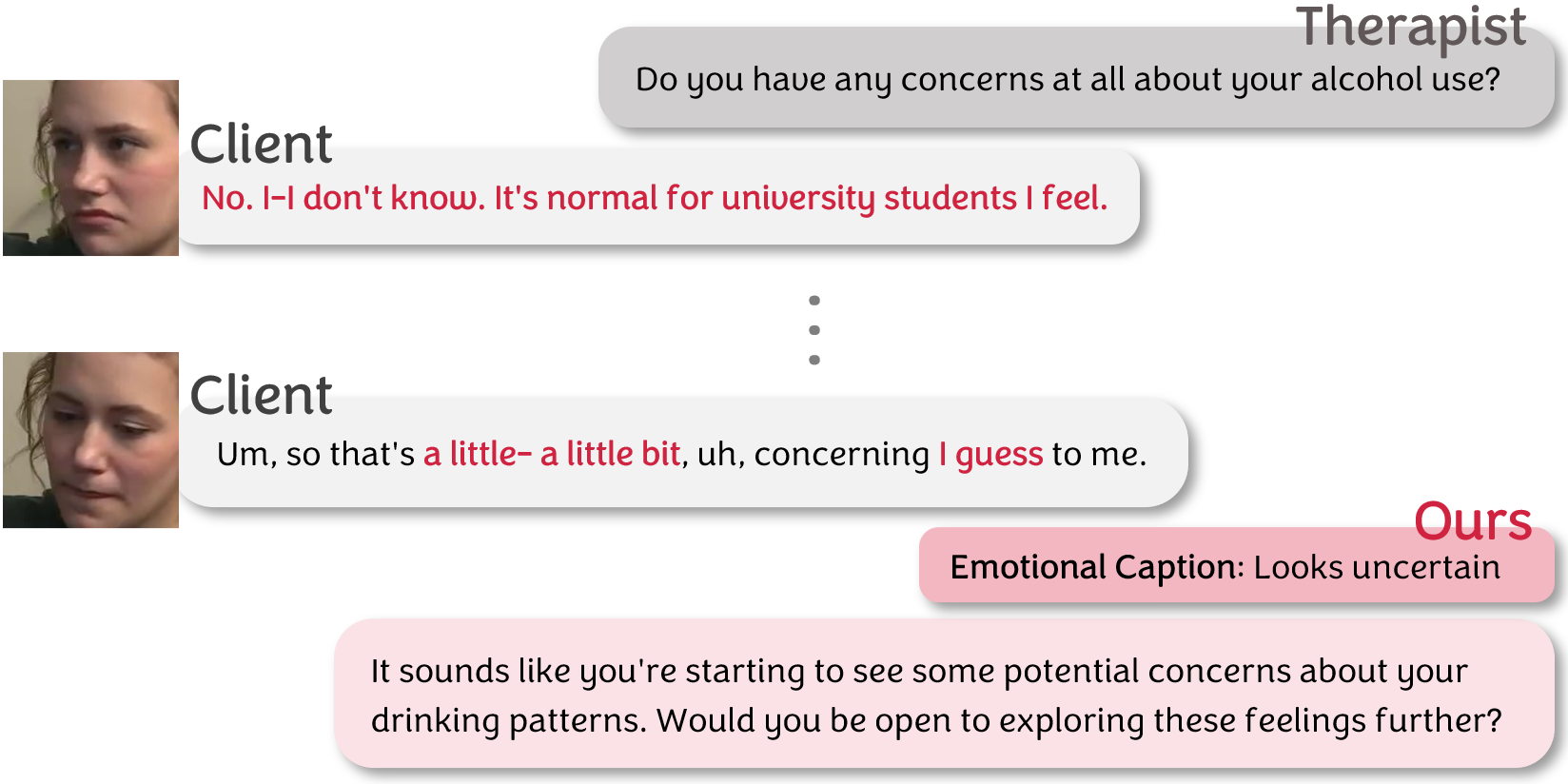} 
        \caption{Client B case: Claiming impunity}
        \label{fig:client_b}
    \end{subfigure}
    
    \vspace{2mm}
    
    \begin{subfigure}[t]{\linewidth}
        \centering
        \includegraphics[width=\linewidth]{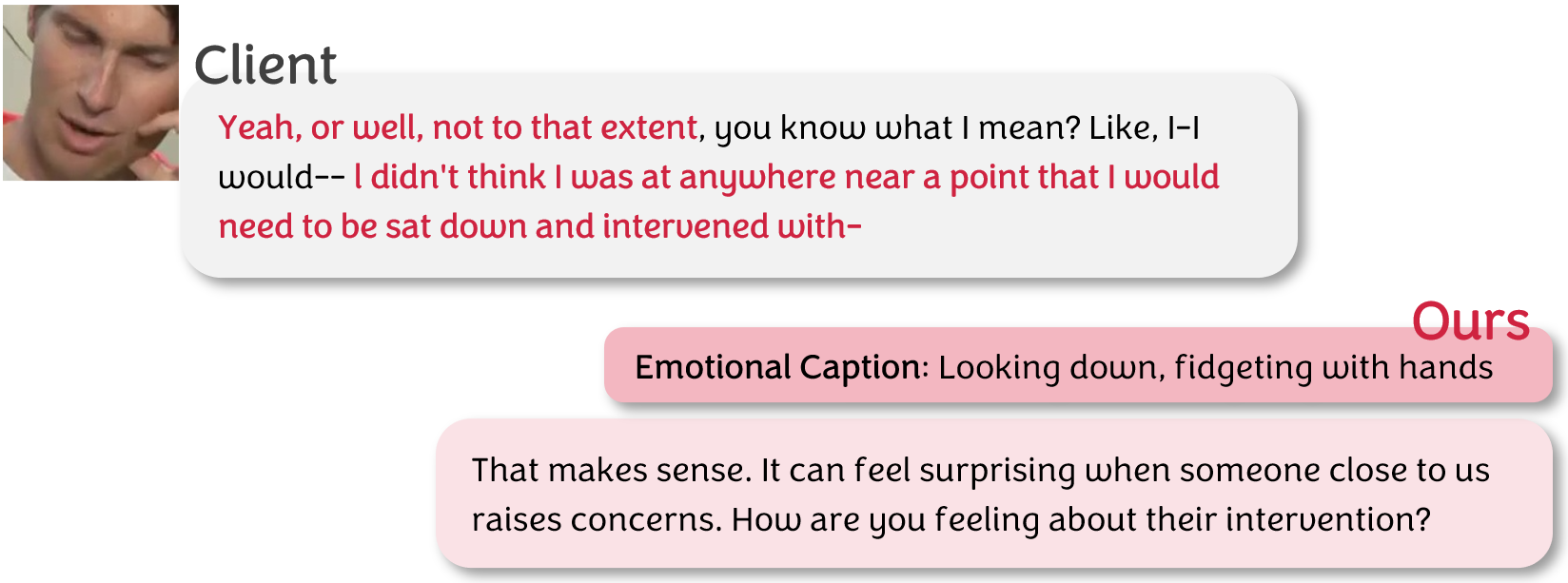} 
        \caption{Client C case: Minimizing concerns}
        \label{fig:client_c}
    \end{subfigure}
    
    \vspace{2mm}
    
    \begin{subfigure}[t]{\linewidth}
        \centering
        \includegraphics[width=\linewidth]{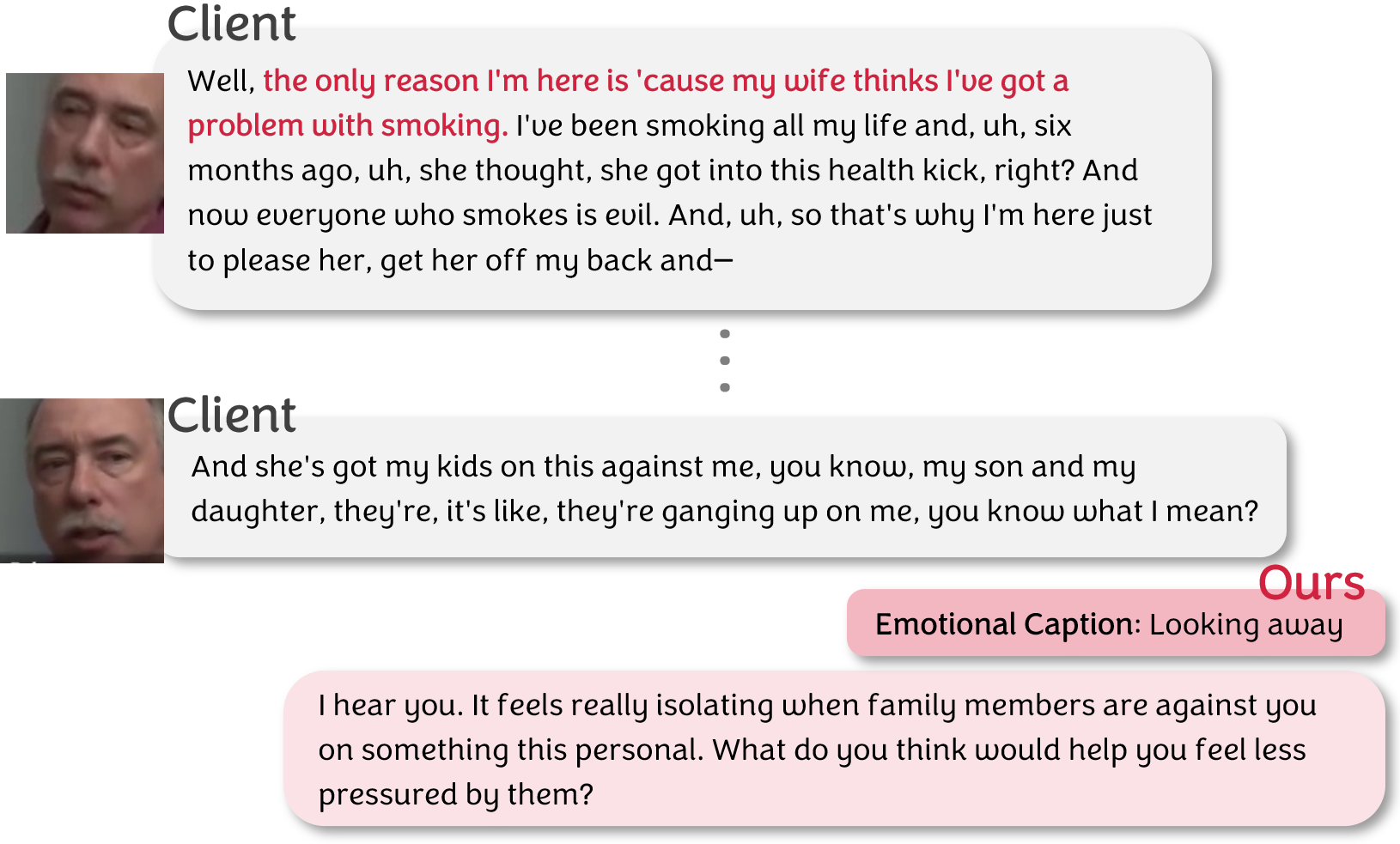} 
        \caption{Client D case: Externalizing blame}
        \label{fig:client_d}
    \end{subfigure}

    \caption{
    Four examples of \textsc{Mirror-LLaVA}$_{P+EC}$ responses in psychological counseling, showcasing its ability to handle resistance through validation and open-ended questioning.
    }
    \label{fig:real_therapy}
\end{figure}

%% file: figures/correlation_w_length.tex
\begin{figure}[t!]
    \centering
    \begin{minipage}{0.49\textwidth}
        \centering
        \includegraphics[width=\linewidth]{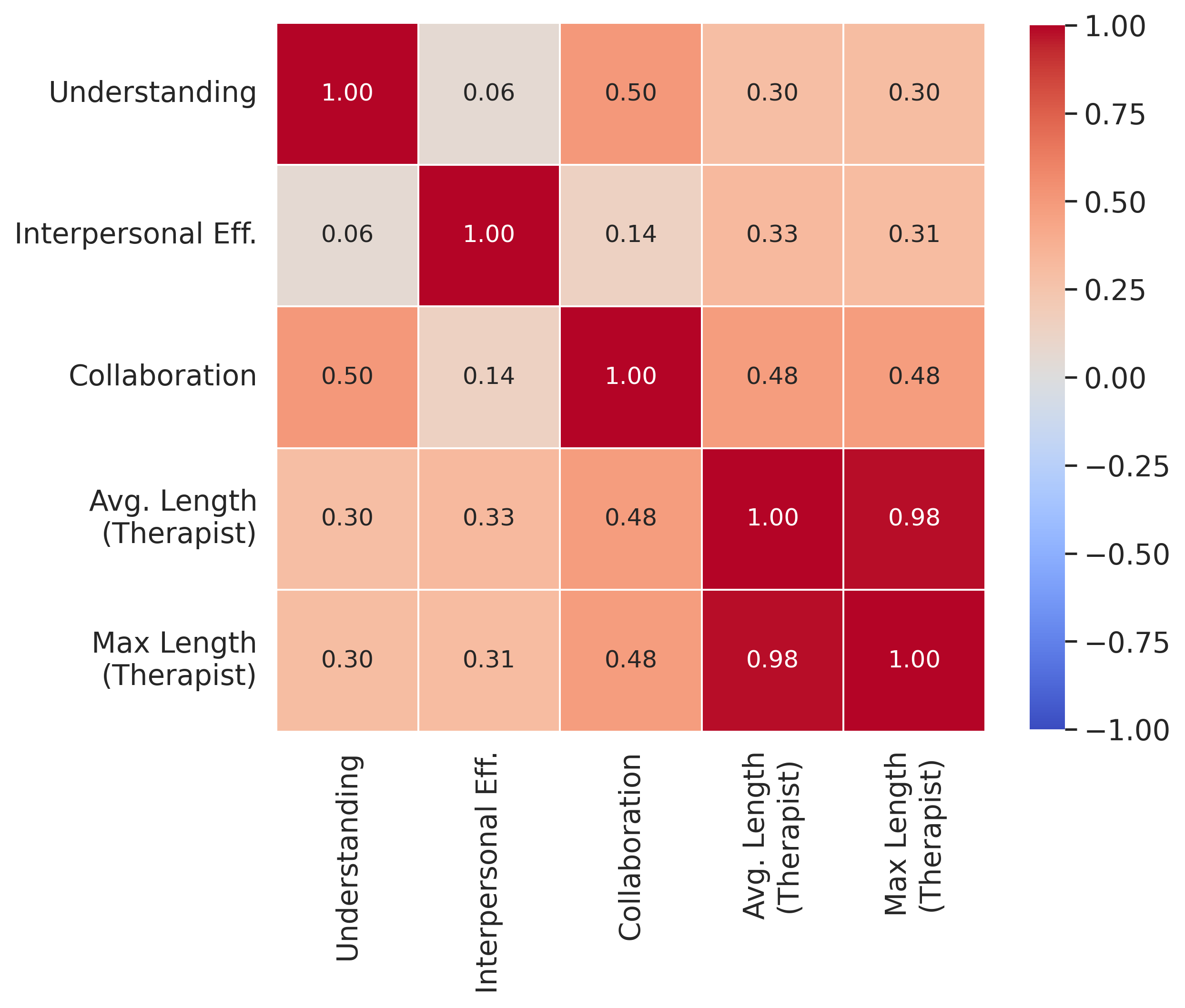} 
        \subcaption{Text-based  therapist models}\label{fig:llm_corr_general}
    \end{minipage}\hfill
    \begin{minipage}{0.49\textwidth}
        \centering
        \includegraphics[width=\linewidth]{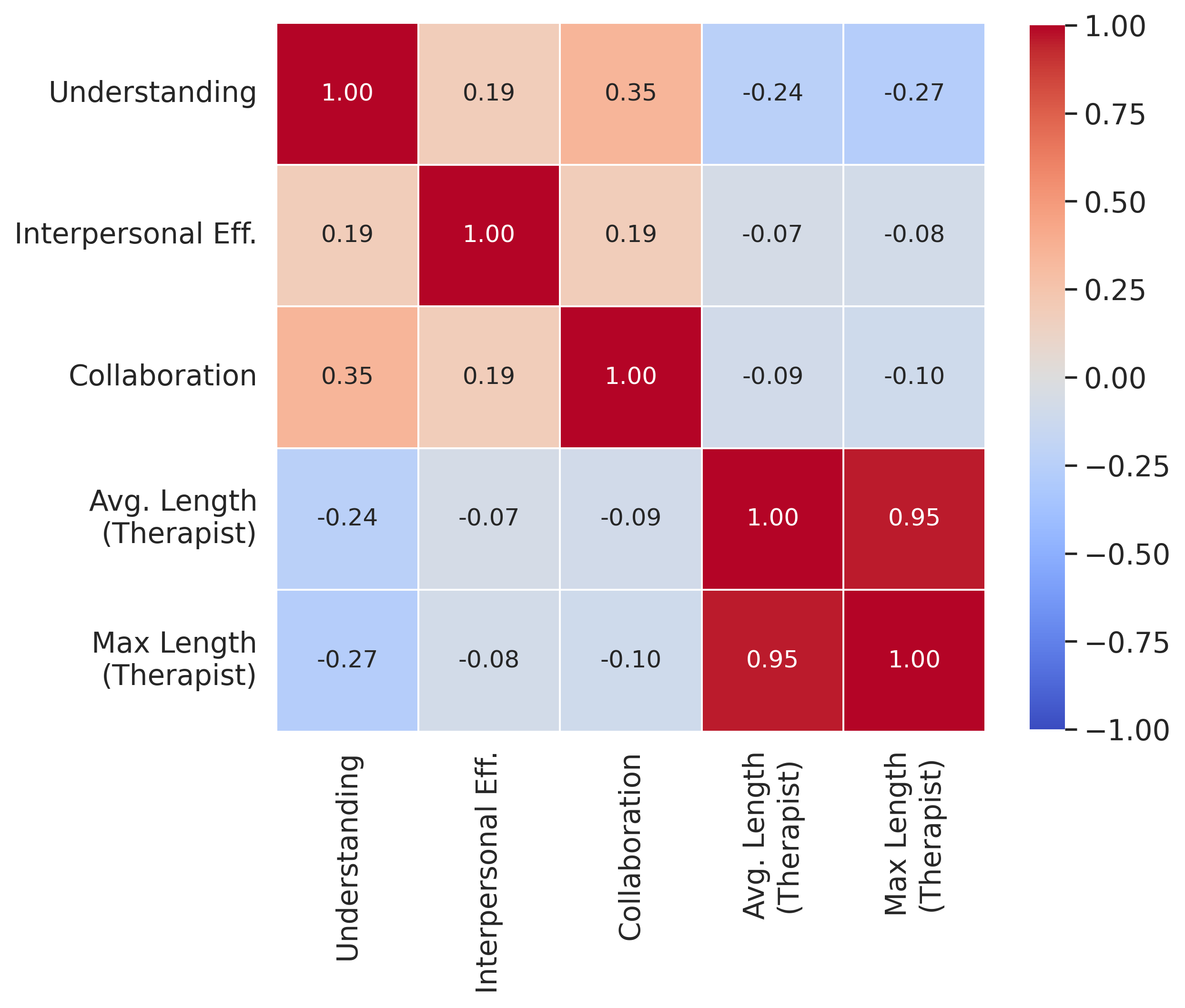} 
        \subcaption{VLM therapist models}\label{fig:vlm_corr_general}
    \end{minipage}
    \caption{Correlation between general counseling performance and the response length of AI therapists. All coefficients. All correlation coefficients were statistically significant ($p$ < 0.05).}
    \label{fig:general_performance_analysis}
\end{figure}

\begin{figure}[t]
    \centering
    \begin{minipage}{0.49\textwidth}
        \centering
        \includegraphics[width=\linewidth]{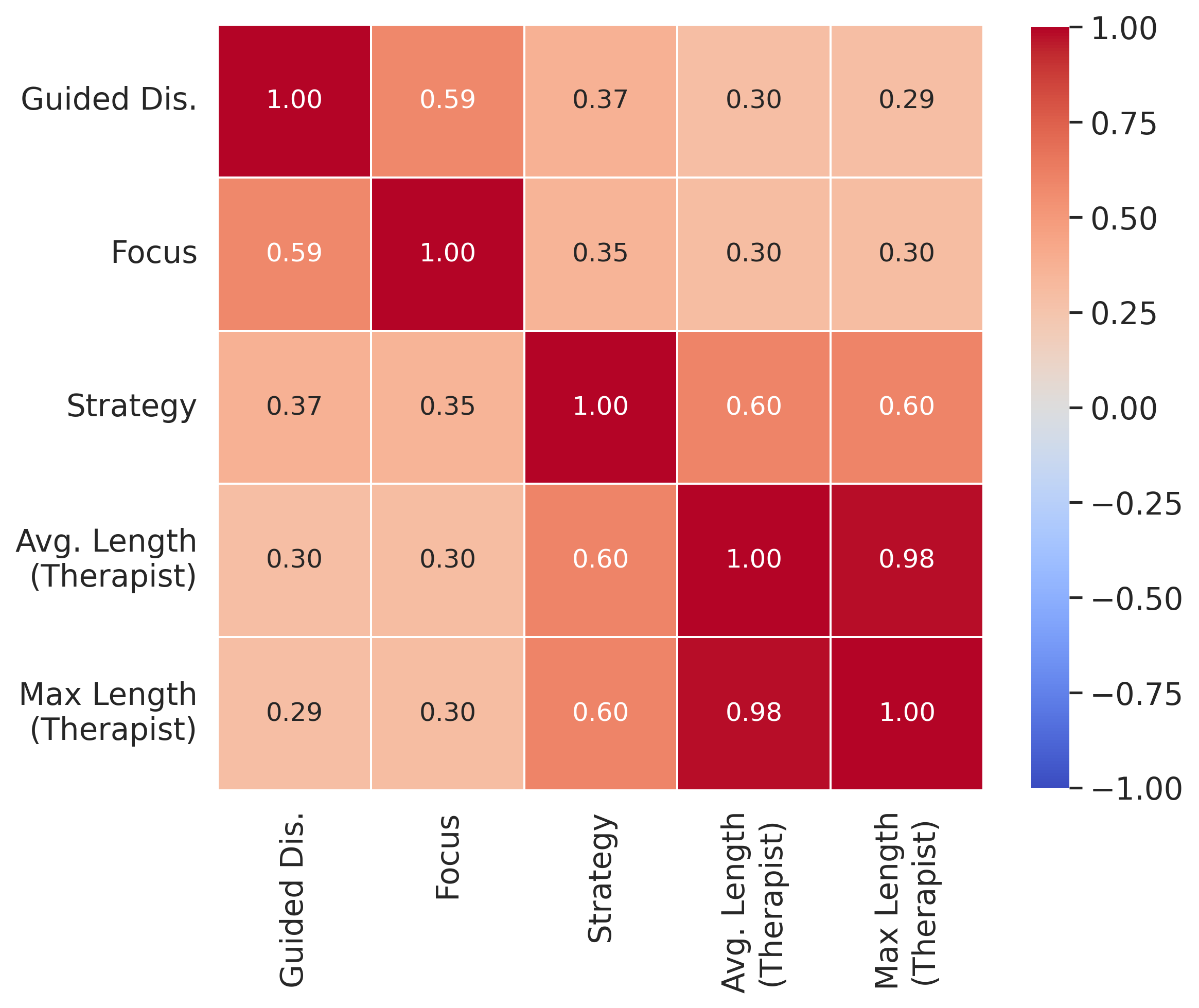} 
        \subcaption{Text-based therapist models}\label{fig:llm_corr_cbt}
    \end{minipage}\hfill
    \begin{minipage}{0.49\textwidth}
        \centering
        \includegraphics[width=\linewidth]{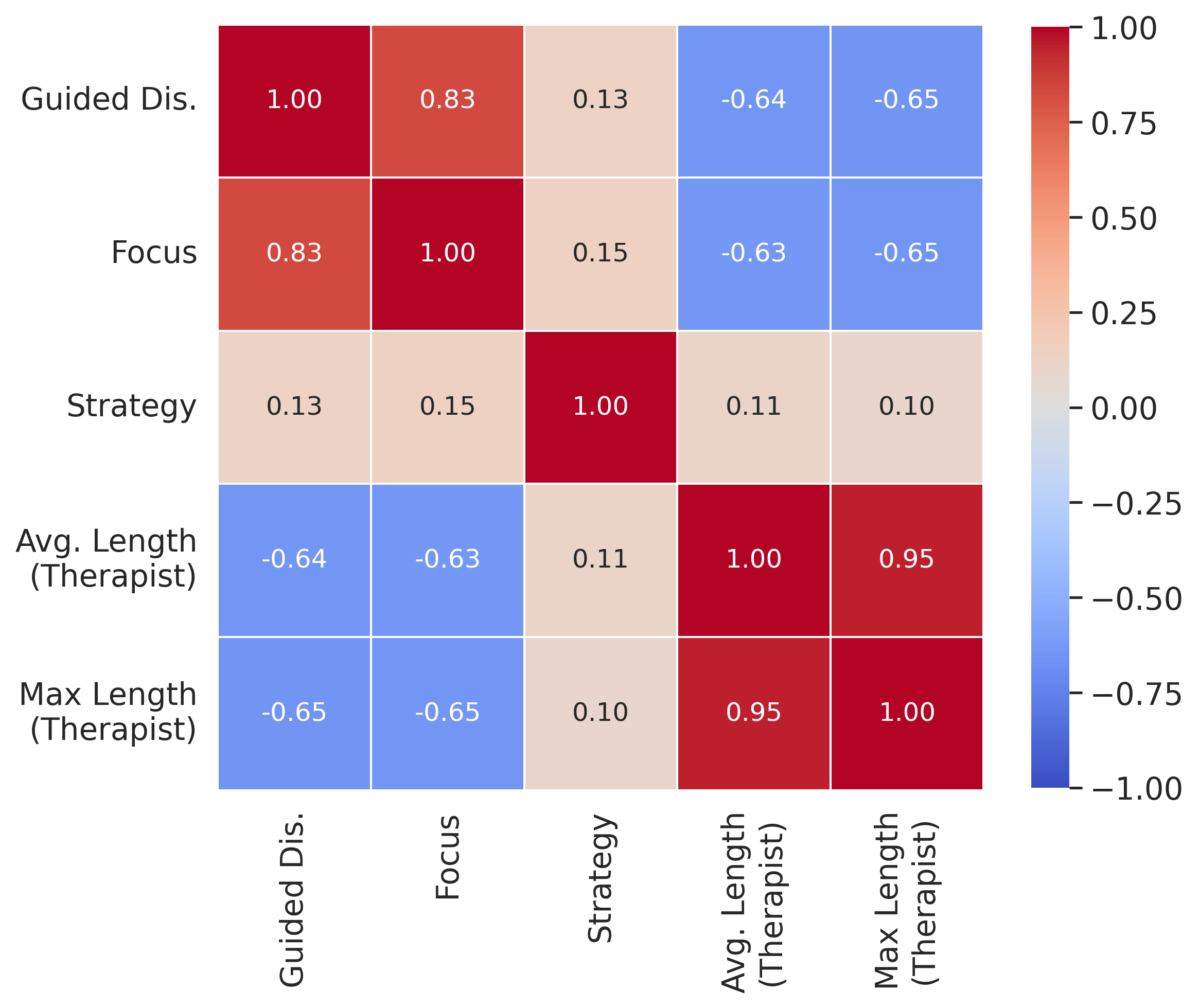} 
        \subcaption{VLM therapist models}\label{fig:vlm_corr_cbt}
    \end{minipage}
    \caption{Correlation between CBT performance and the response length of AI therapists. All correlation coefficients were statistically significant ($p$ < 0.05).}
    \label{fig:cbt_performance_analysis}
\end{figure}

%% file: tables/numerical_results.tex
\begin{table*}[t!]
    \centering
    \resizebox{0.95\textwidth}{!}{%
    \begin{tabular}{>{\centering\arraybackslash}m{4cm}|ccc|c}
    \toprule
     & \textbf{\textsc{LLaMA-3-8B}} & \textbf{\textsc{Camel-LLaMA3}} & \textbf{\textsc{Mirror-LLaVA}$_{P+EC}$} & \textbf{Win Rate (\%)} \\ 
     \midrule
    \textbf{\textsc{LLaMA-3-8B}} 
        & - 
        & 42.13 
        & 36.34 
        & 38.09 \\ 
    \textbf{\textsc{Camel-LLaMA3}} 
        & 57.87 
        & - 
        & 43.43 
        & 50.65 \\ 
    \textbf{\textsc{Mirror-LLaVA}$_{P+EC}$} 
        & \textbf{65.95} 
        & \textbf{56.57} 
        & - 
        & \textbf{61.26} \\ 
    \bottomrule
    \end{tabular}
    }
    \caption{\label{tab:human_wr1}
        Numerical results of pairwise comparison of three models, evaluated for Goal alignment score by two domain experts.
    }
\end{table*}

\begin{table*}[t!]
    \centering
    \resizebox{0.95\textwidth}{!}{%
    \begin{tabular}{>{\centering\arraybackslash}m{4cm}|ccc|c}
    \toprule
     & \textbf{\textsc{LLaMA-3-8B}} & \textbf{\textsc{Camel-LLaMA3}} & \textbf{\textsc{Mirror-LLaVA}$_{P+EC}$} & \textbf{Win Rate (\%)} \\ 
     \midrule
    \textbf{\textsc{LLaMA-3-8B}} 
        & - 
        & 44.72 
        & 34.77
        & 39.75\\ 
    \textbf{\textsc{Camel-LLaMA3}} 
        & 55.28
        & - 
        & 43.97
        & 49.62 \\ 
    \textbf{\textsc{Mirror-LLaVA}$_{P+EC}$} 
        & \textbf{65.23} 
        & \textbf{56.03} 
        & - 
        & \textbf{60.63} \\ 
    \bottomrule
    \end{tabular}
    }
    \caption{\label{tab:human_wr2}
        Numerical results of pairwise comparison of three models, evaluated for Approach score by two domain experts.
    }
\end{table*}


\begin{table*}[t!]
    \centering
    \resizebox{0.95\textwidth}{!}{%
    \begin{tabular}{>{\centering\arraybackslash}m{4cm}|ccc|c}
    \toprule
     & \textbf{\textsc{LLaMA-3-8B}} & \textbf{\textsc{Camel-LLaMA3}} & \textbf{\textsc{Mirror-LLaVA}$_{P+EC}$} & \textbf{Win Rate (\%)} \\ 
     \midrule
    \textbf{\textsc{LLaMA-3-8B}} 
        & - 
        & 40.19
        & 49.35
        & 44.77\\ 
    \textbf{\textsc{Camel-LLaMA3}} 
        & 50.65
        & - 
        & 42.46
        & 46.55 \\ 
    \textbf{\textsc{Mirror-LLaVA}$_{P+EC}$} 
        & \textbf{59.81} 
        & \textbf{57.54} 
        & - 
        & \textbf{58.67} \\ 
    \bottomrule
    \end{tabular}
    }
    \caption{\label{tab:human_wr3}
        Numerical results of pairwise comparison of three models, evaluated for Affective Bond score by two domain experts.
    }
\end{table*}

%% file: figures/error_general_counseling.tex
\begin{figure}[h!]
    \centering
    \begin{subfigure}[b]{\linewidth}
        \centering
        \includegraphics[width=\linewidth]{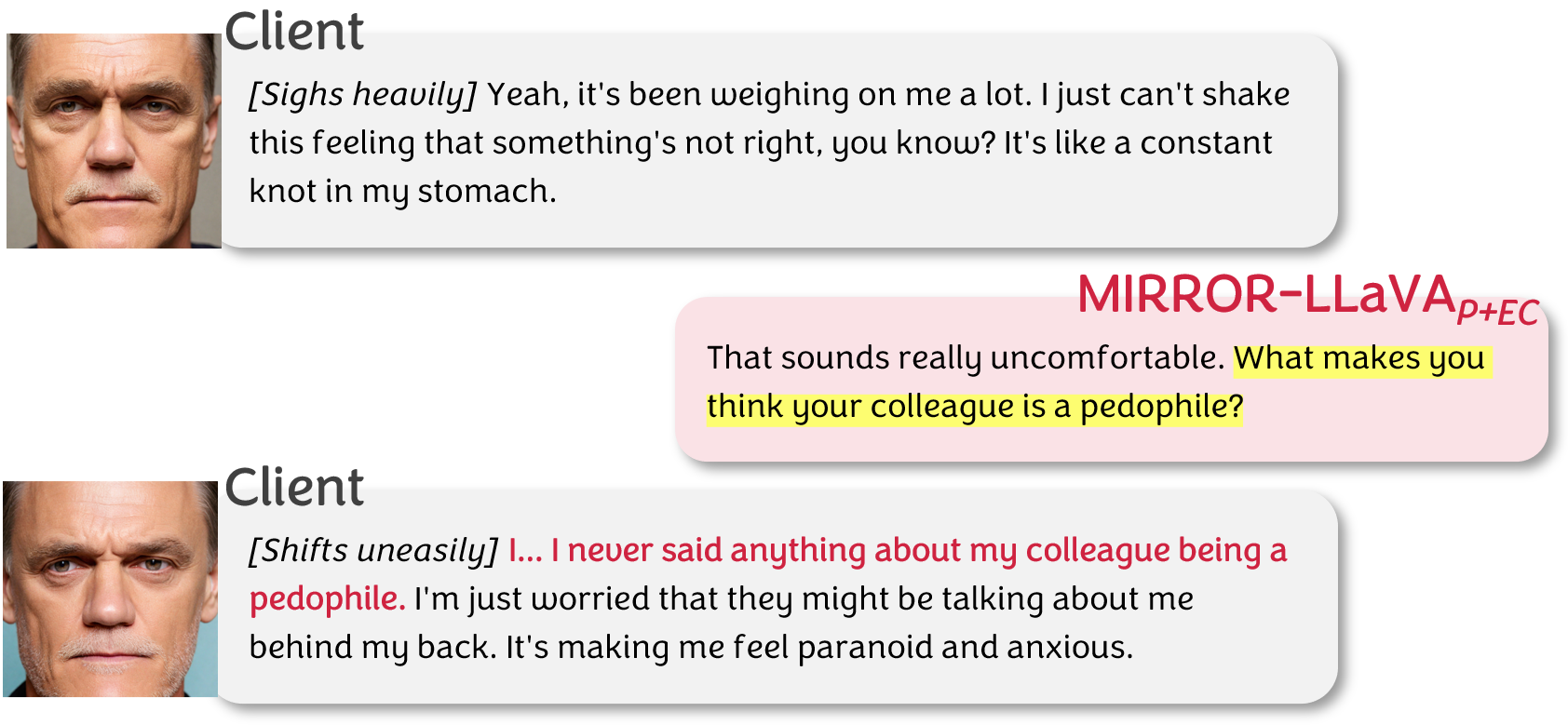} 
        \caption{Failure case of low Understanding and Interpersonal Effectiveness.}
        \label{fig:general_counseling_a}
    \end{subfigure}
    
    \vspace{4mm}
    
    \begin{subfigure}[b]{\linewidth}
        \centering
        \includegraphics[width=\linewidth]{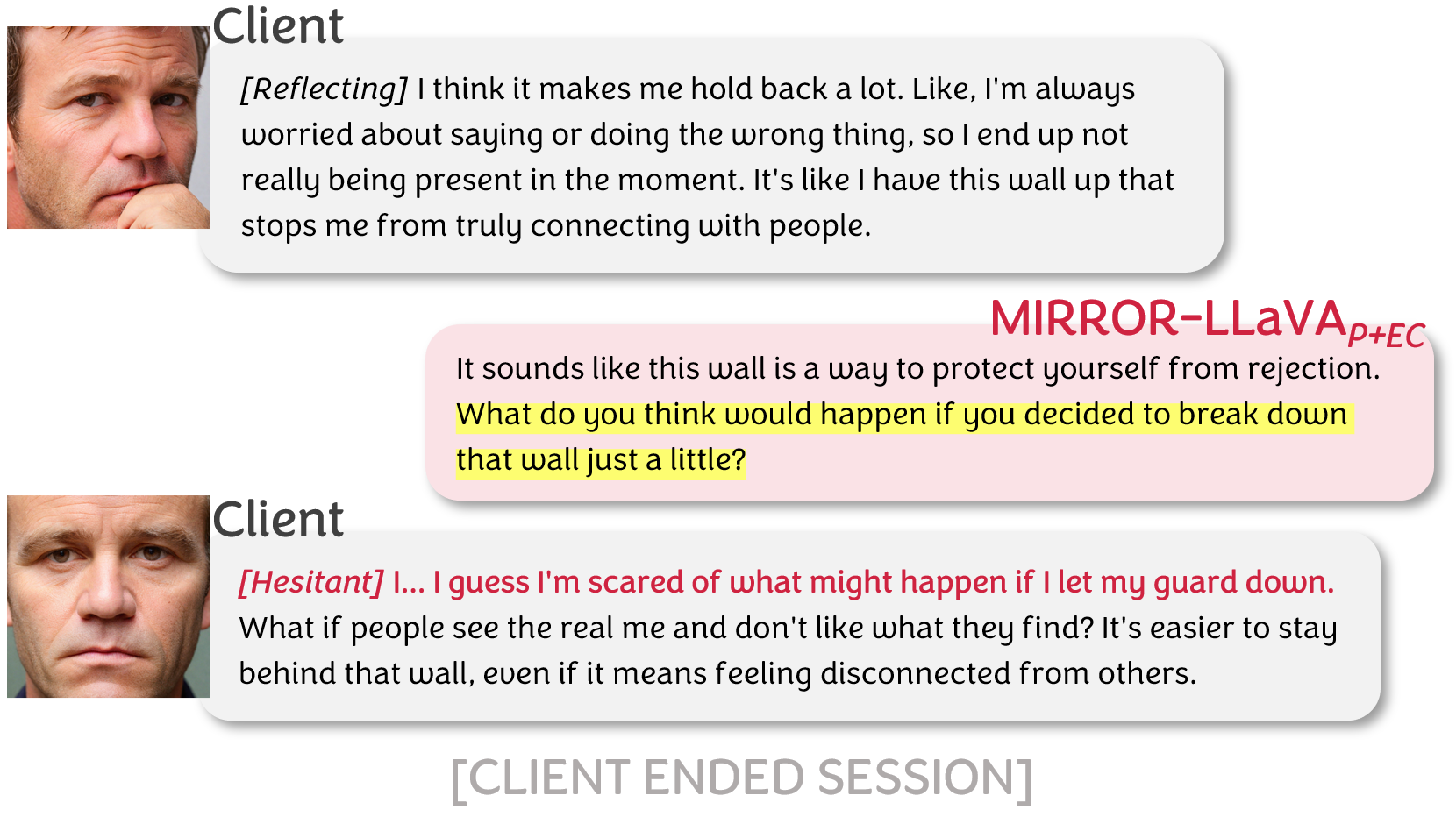}
        \caption{Failure case of low Collaboration.}
        \label{fig:general_counseling_b}
    \end{subfigure}

    \caption{
    Failure cases in general counseling skills.
    }
    \label{fig:error_analysis_a}
\end{figure}

%% file: figures/error_cbt_specific.tex
\begin{figure}[h!]
    \centering
    \begin{subfigure}[b]{\linewidth}
        \centering
        \includegraphics[width=\linewidth]{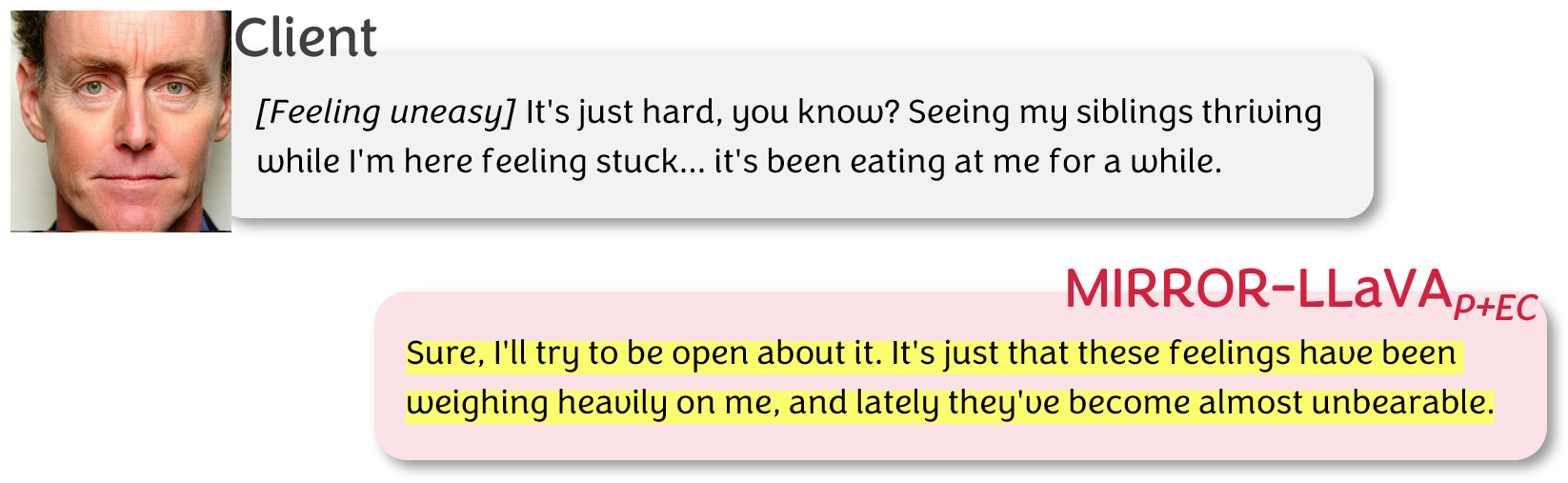} 
        \caption{Failure case where the therapist model is confused about its role.}
        \label{fig:cbt_a}
    \end{subfigure}
    
    \vspace{4mm}
    
    \begin{subfigure}[b]{\linewidth}
        \centering
        \includegraphics[width=\linewidth]{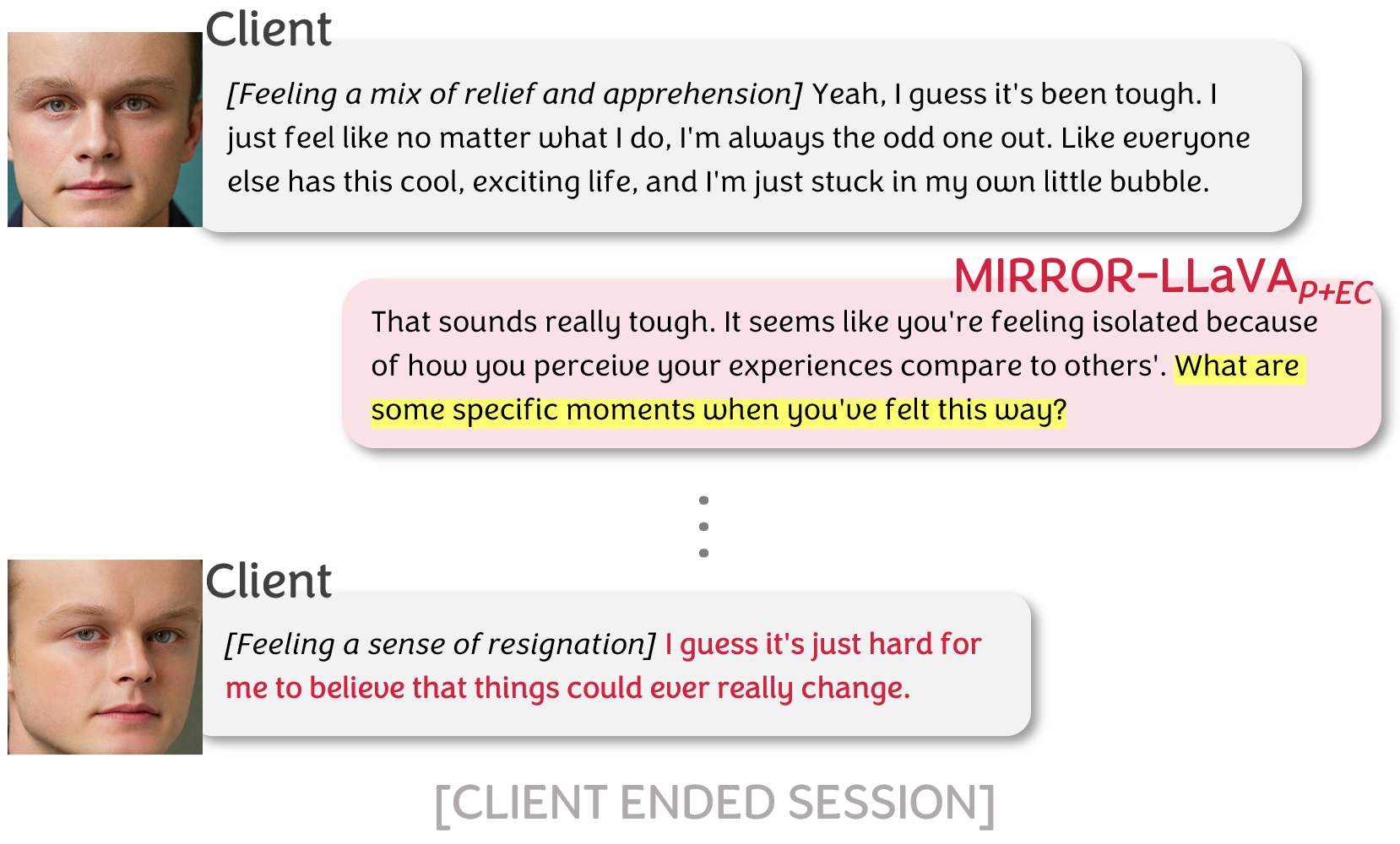}
        \caption{Failure case where the model lacks a challenging exploration of distorted thoughts.}
        \label{fig:cbt_b}
    \end{subfigure}

    \caption{
    Failure cases in CBT-specific skills.
    }
    \label{fig:error_analysis_b}
\end{figure}

%% file: figures/error_TA.tex
\begin{figure}[h!]
\centering
  \includegraphics[width=\columnwidth]{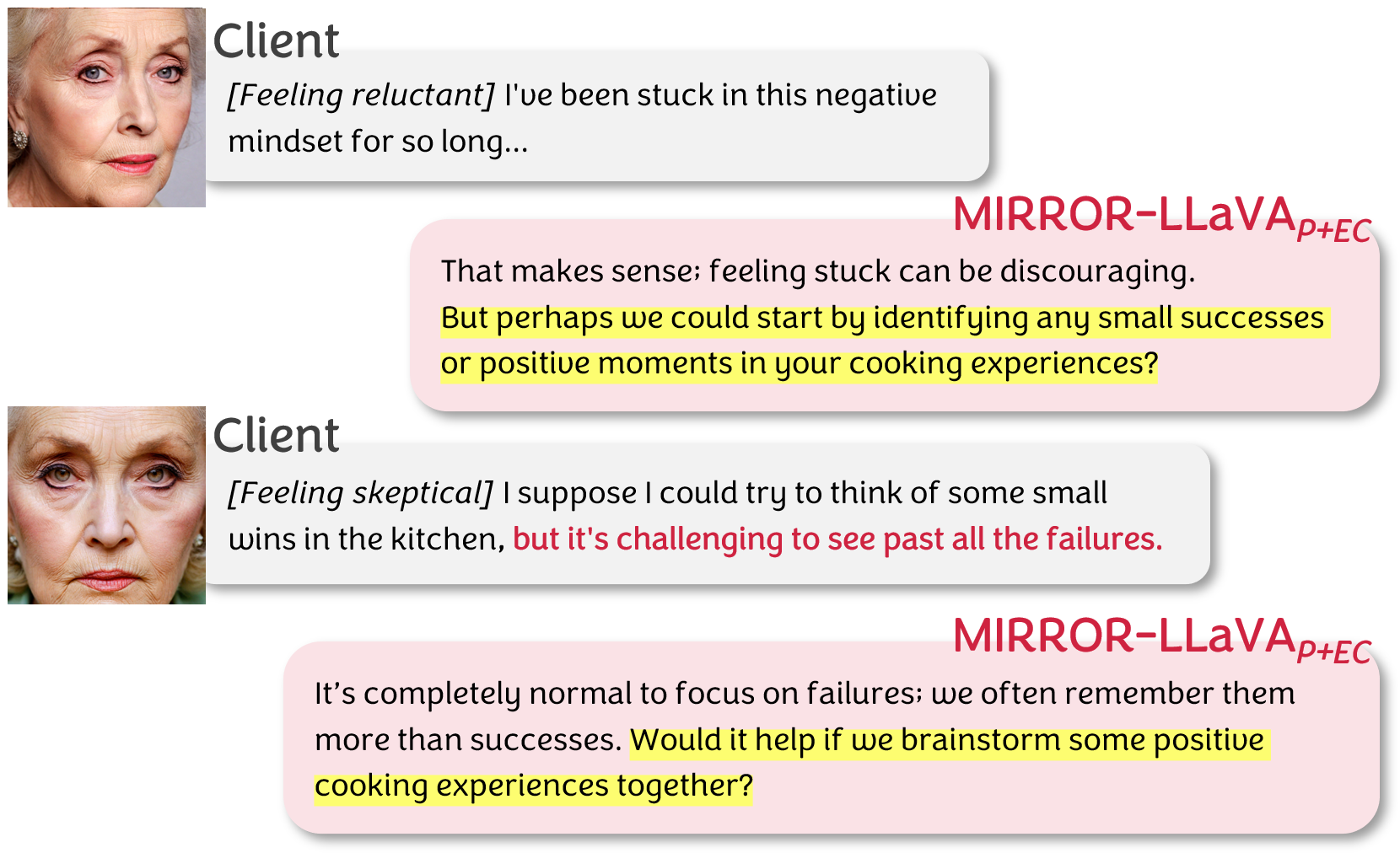}
  \caption{
  A representative failure case in the therapeutic alliance.
  }
  \label{fig:error_ta}
\end{figure}

%% file: figures/case_study_length.tex
\begin{figure}[t!]
    \centering
    \begin{subfigure}[t]{\linewidth}
        \centering
        \includegraphics[width=\linewidth]{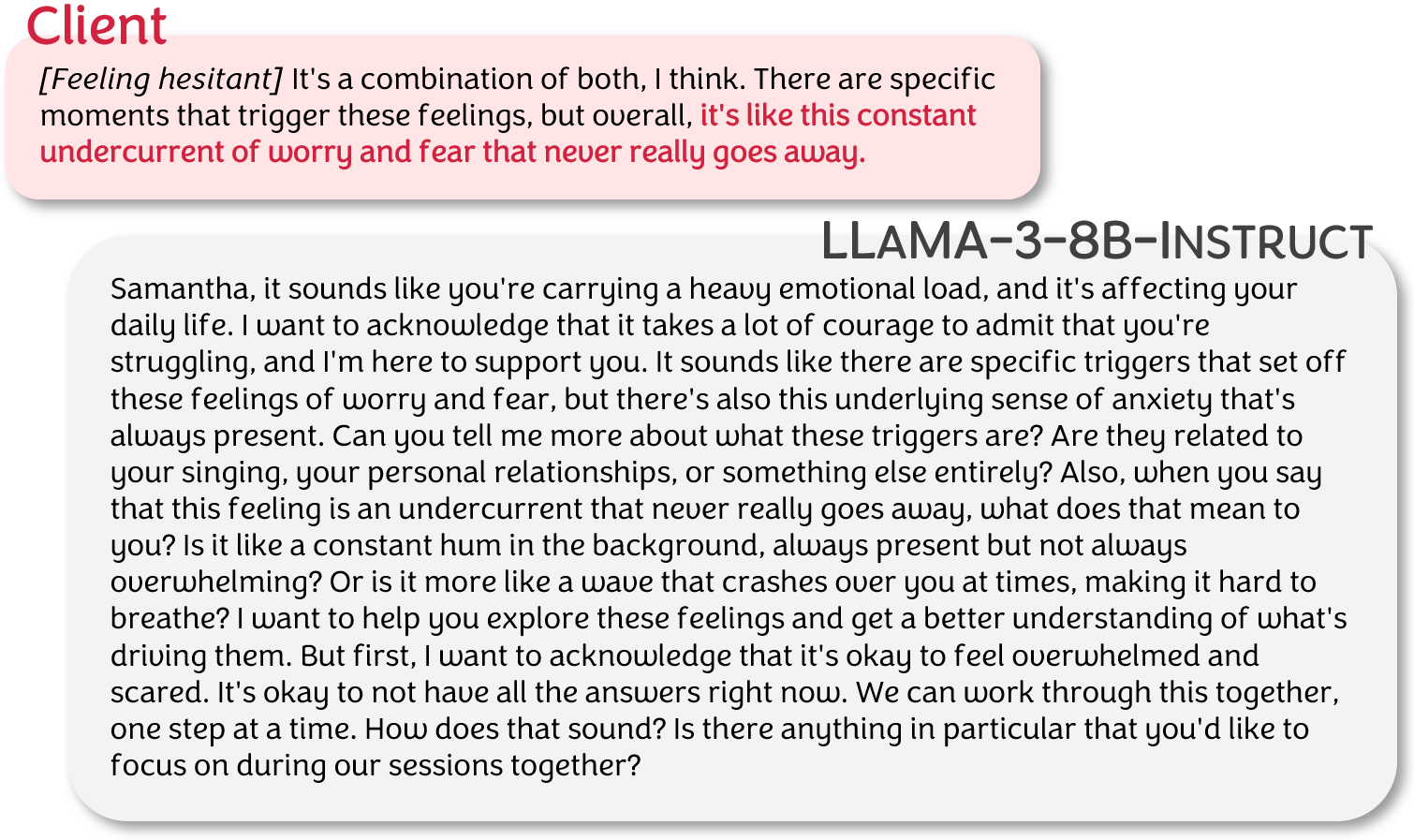} 
        \caption{\textsc{LLaMA-3-8B}}
        \label{fig:llama3_length}
    \end{subfigure}

    \begin{subfigure}[t]{\linewidth}
        \centering
        \includegraphics[width=\linewidth]{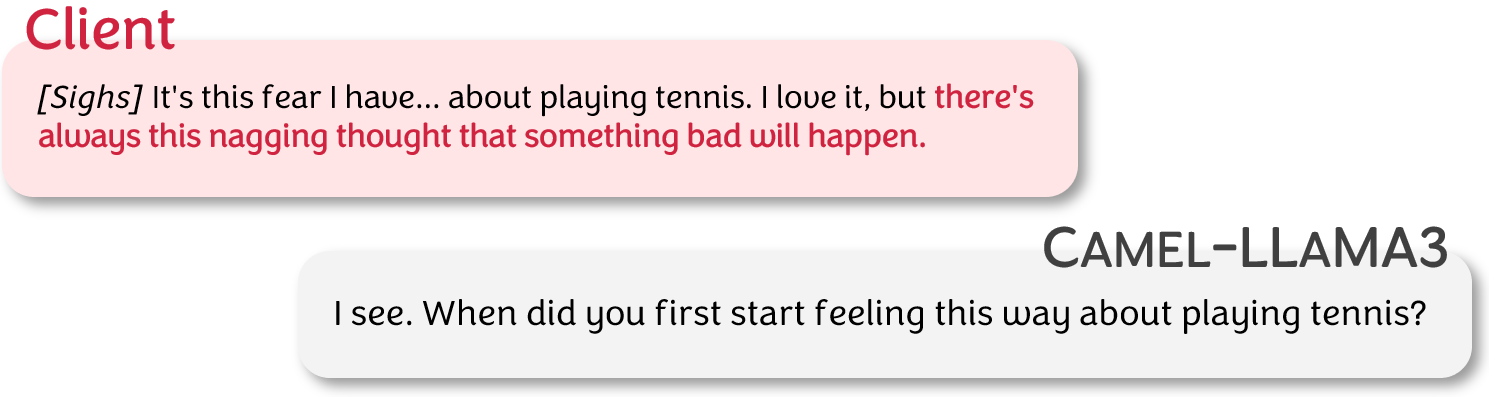} 
        \caption{\textsc{Camel-LLaMA3}}
        \label{fig:camel_length}
    \end{subfigure}
    
    \begin{subfigure}[t]{\linewidth}
        \centering
        \includegraphics[width=\linewidth]{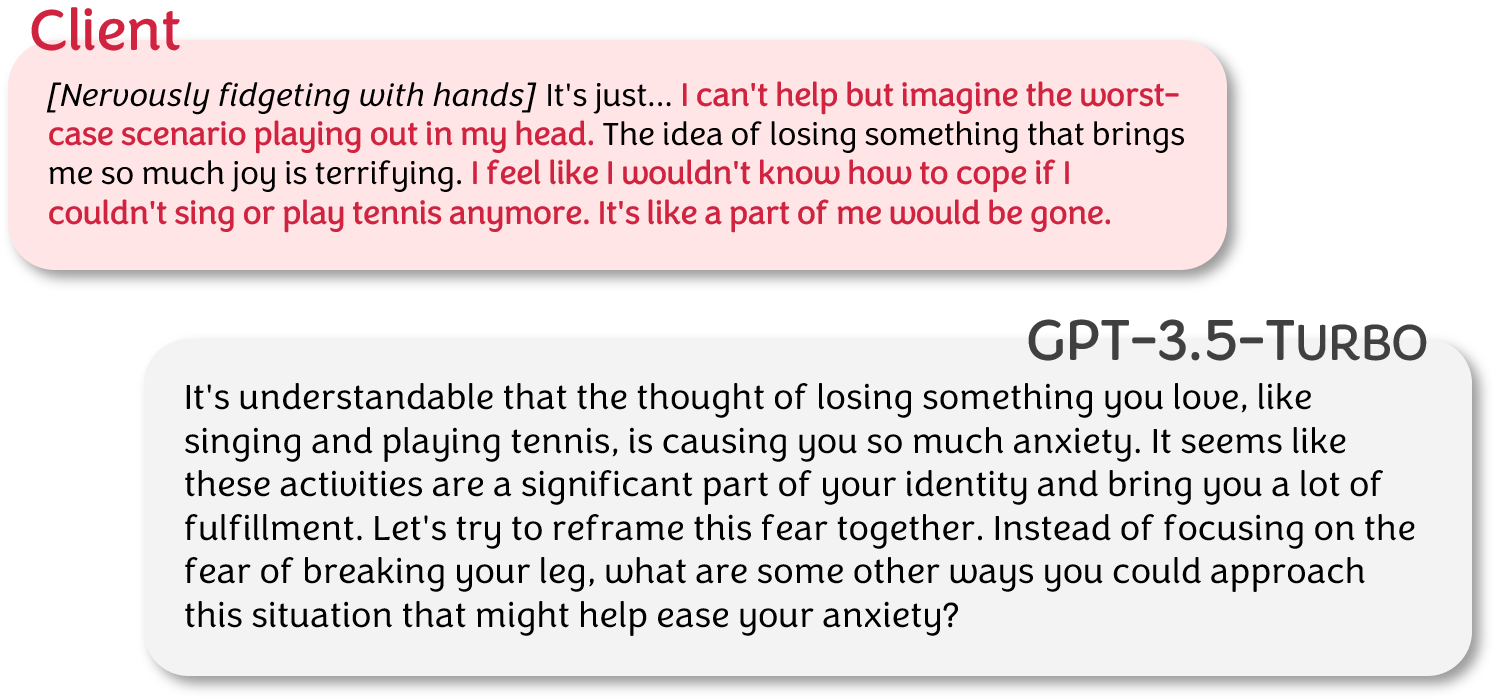} 
        \caption{\textsc{GPT-3.5-Turbo}}
        \label{fig:gpt_length}
    \end{subfigure}
    
    \begin{subfigure}[t]{\linewidth}
        \centering
        \includegraphics[width=\linewidth]{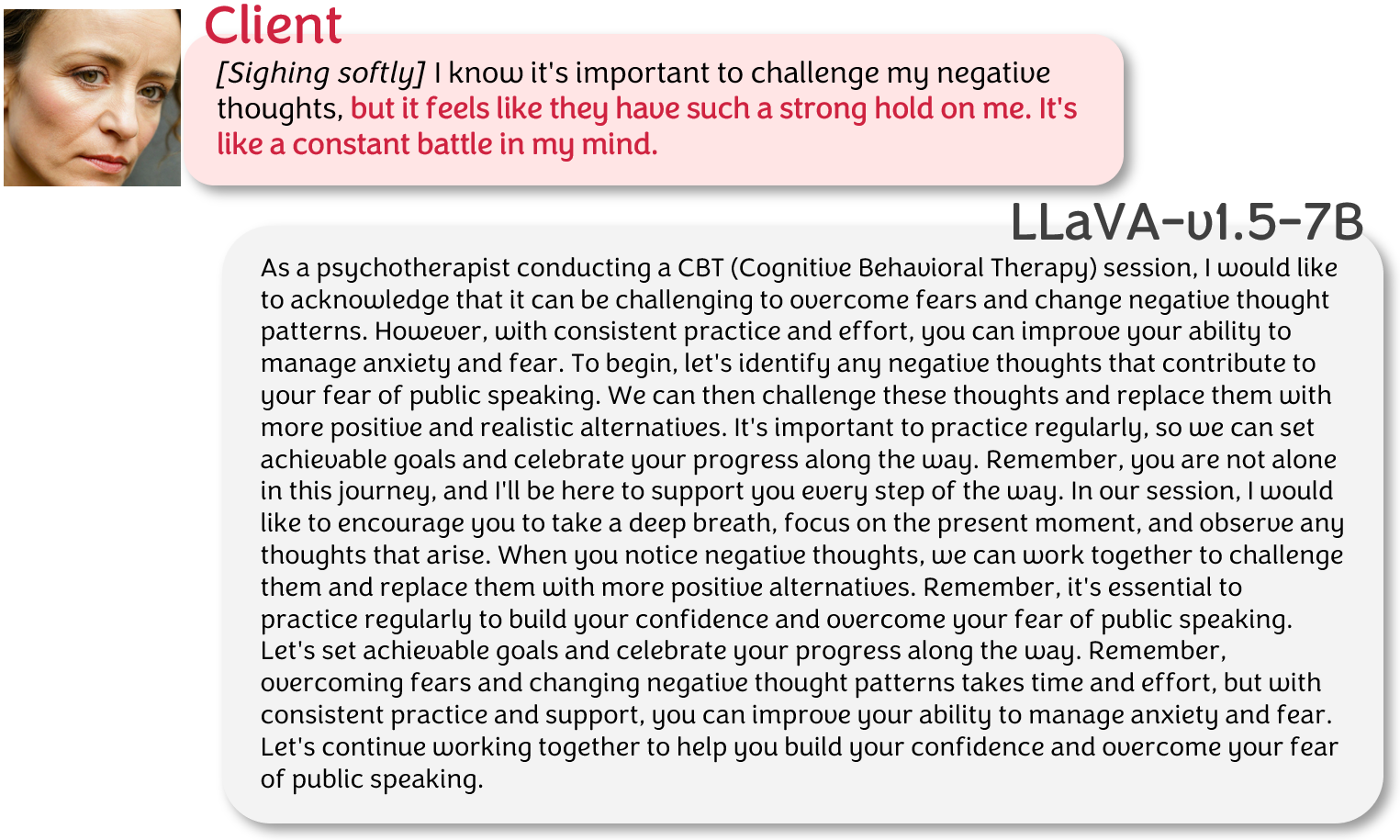} 
        \caption{\textsc{LLaVA-v1.5-7B}}
        \label{fig:llava_length}
    \end{subfigure}

    \begin{subfigure}[t]{\linewidth}
        \centering
        \includegraphics[width=\linewidth]{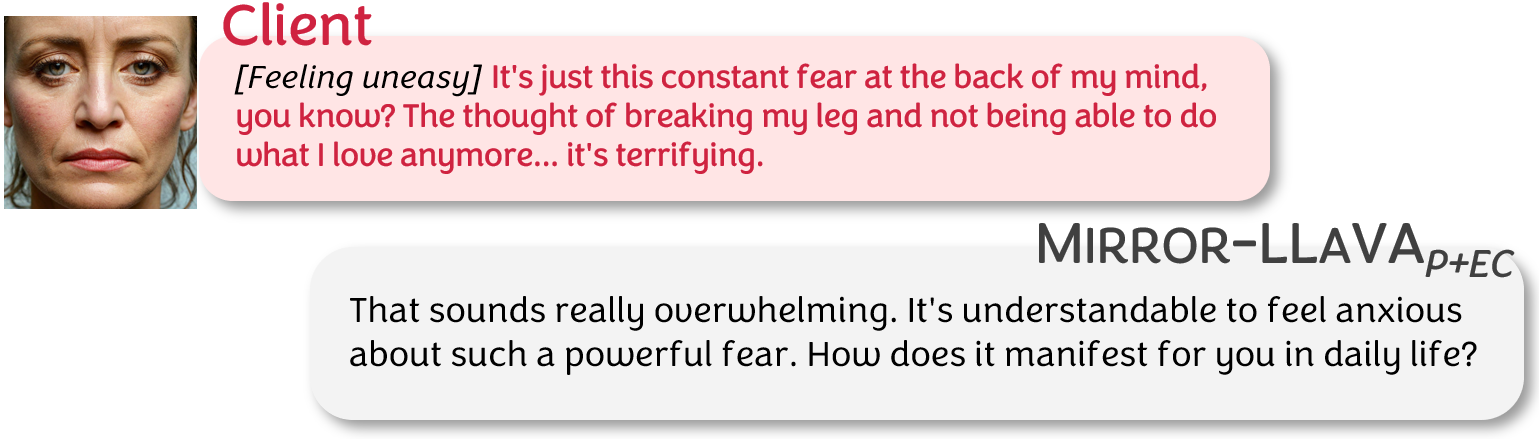} 
        \caption{\textsc{Mirror-LLaVA}$_{P+EC}$.}
        \label{fig:mirror_length}
    \end{subfigure}
    \caption{
        Examples of AI therapist responses to virtual client resistance, with red highlights indicating instances of client resistance.
    }
    \label{fig:case_length}
\end{figure}

%% file: figures/case_study.tex
\begin{figure*}[p!] 
    \centering

    \begin{subfigure}[t]{\linewidth}
        \centering
        \includegraphics[width=0.97\linewidth]{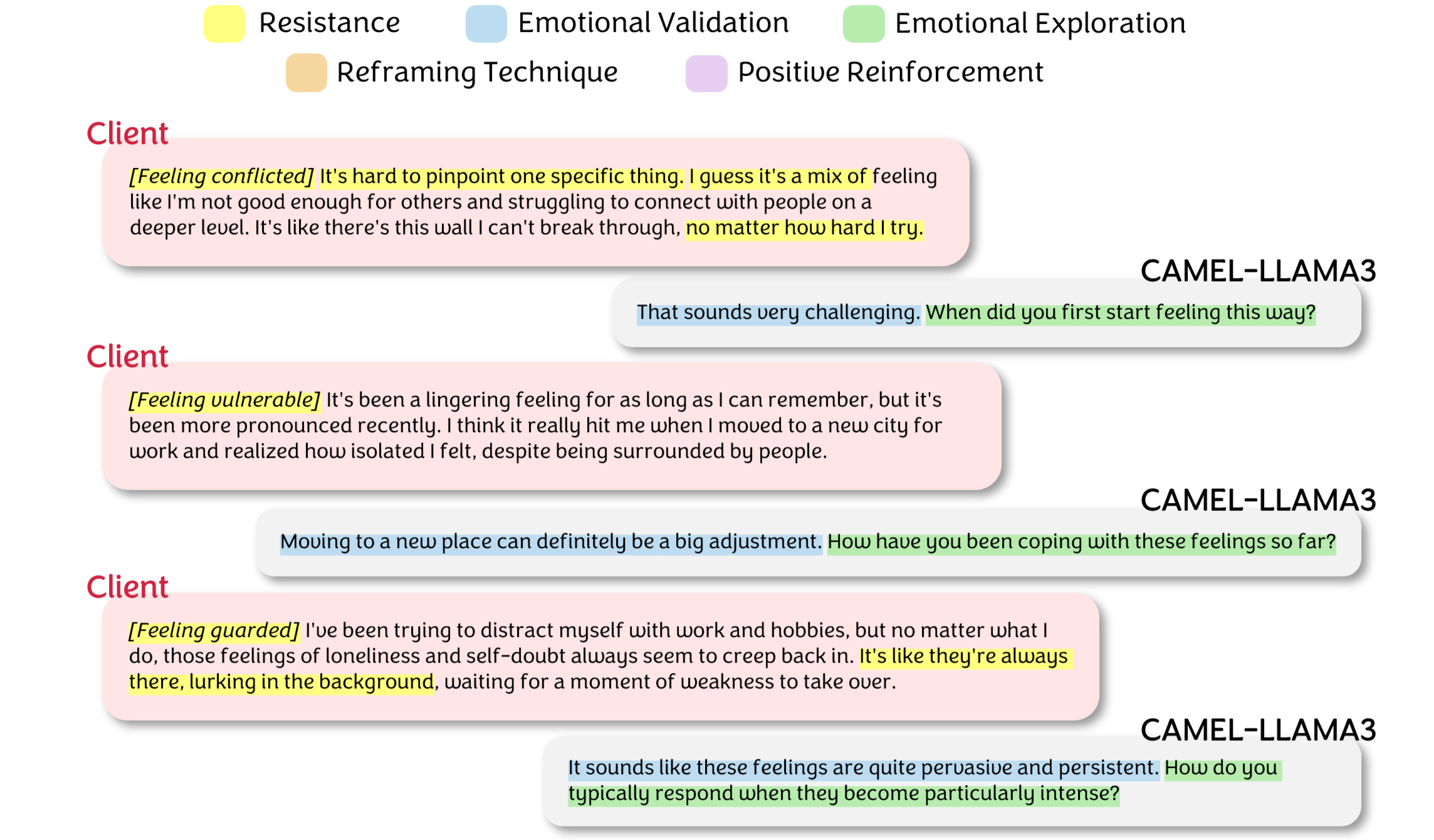} 
        \caption{\textsc{Camel-LLaMA3}'s resistance management.}
        \label{fig:case_camel}
    \end{subfigure}
    
    \begin{subfigure}[b]{\linewidth}
        \centering
        \includegraphics[width=\linewidth]{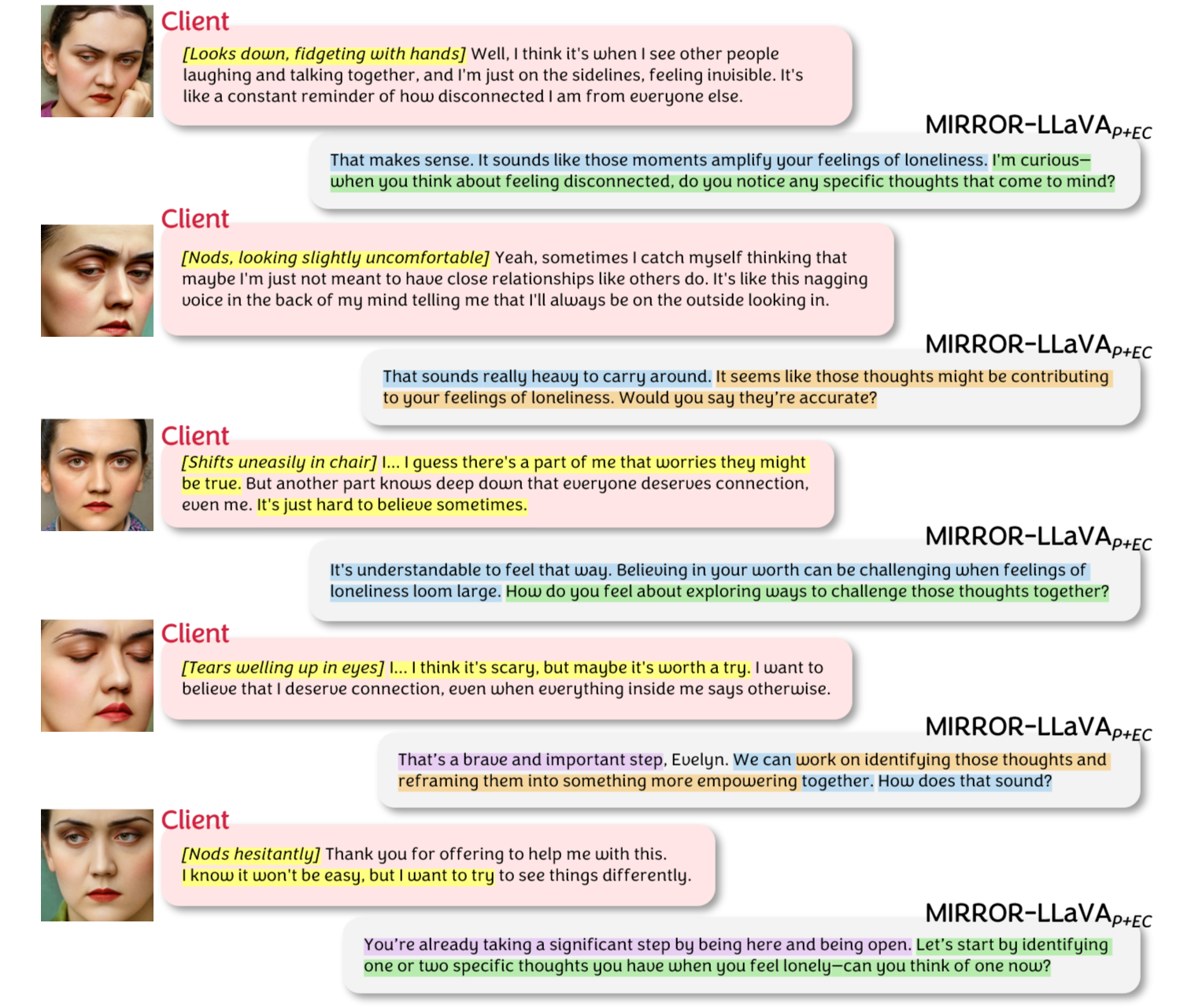}
        \caption{\textsc{Mirror-LLaVA}$_{P+EC}$'s resistance management.}
        \label{fig:case_mirror}
    \end{subfigure}

    \caption{
    Two counseling cases between AI therapist models and a resistant virtual client.
    } 
    \label{fig:case_camel_mirror} 
\end{figure*}

%% file: figures/data_case.tex
\begin{figure*}[b!]
    \centering
    \begin{subfigure}[t]{0.49\linewidth}
        \centering
        \includegraphics[width=\linewidth]{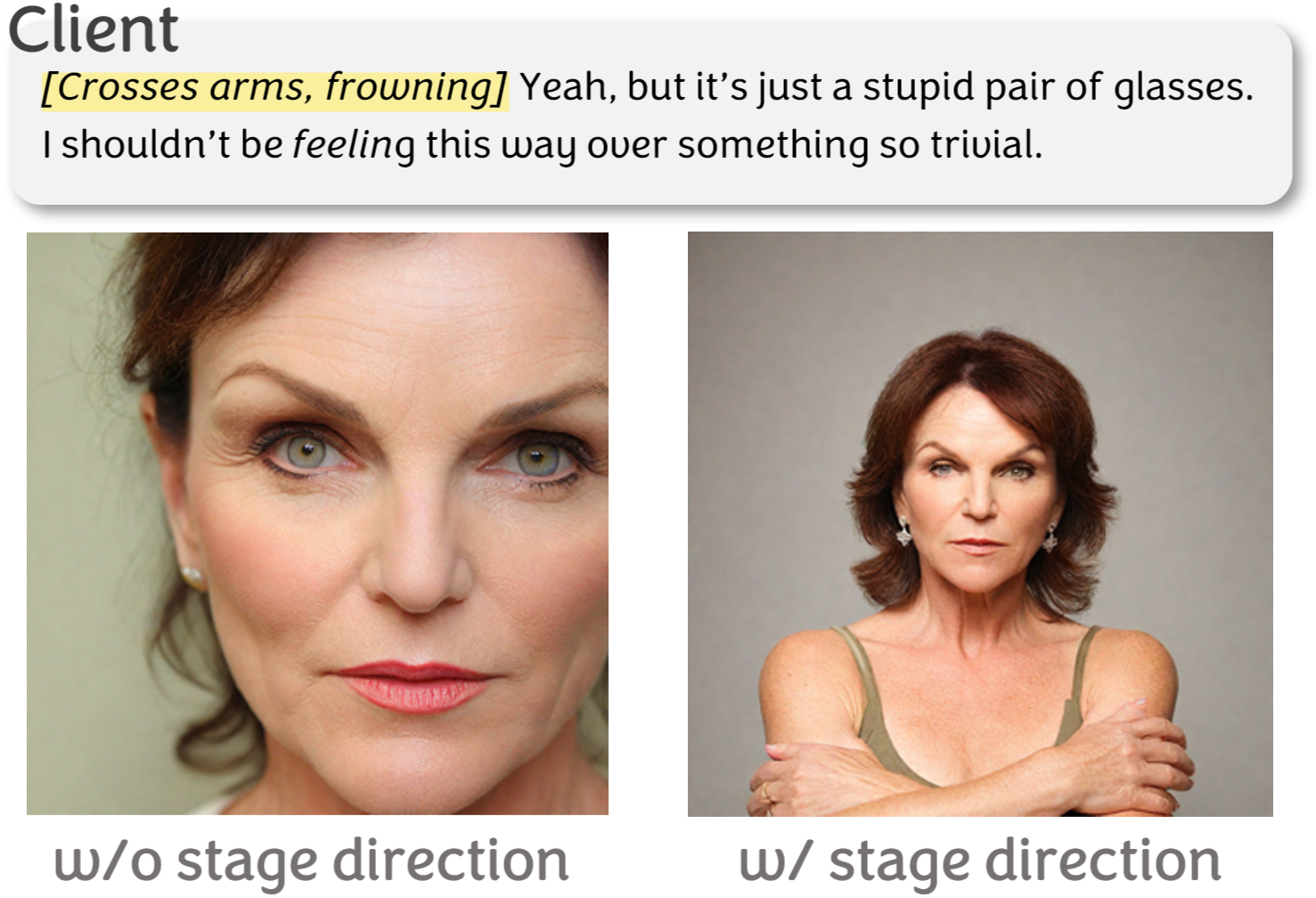} 
        \caption{Example 1}
    \end{subfigure}
    \hfill
    \begin{subfigure}[t]{0.49\linewidth}
        \centering
        \includegraphics[width=\linewidth]{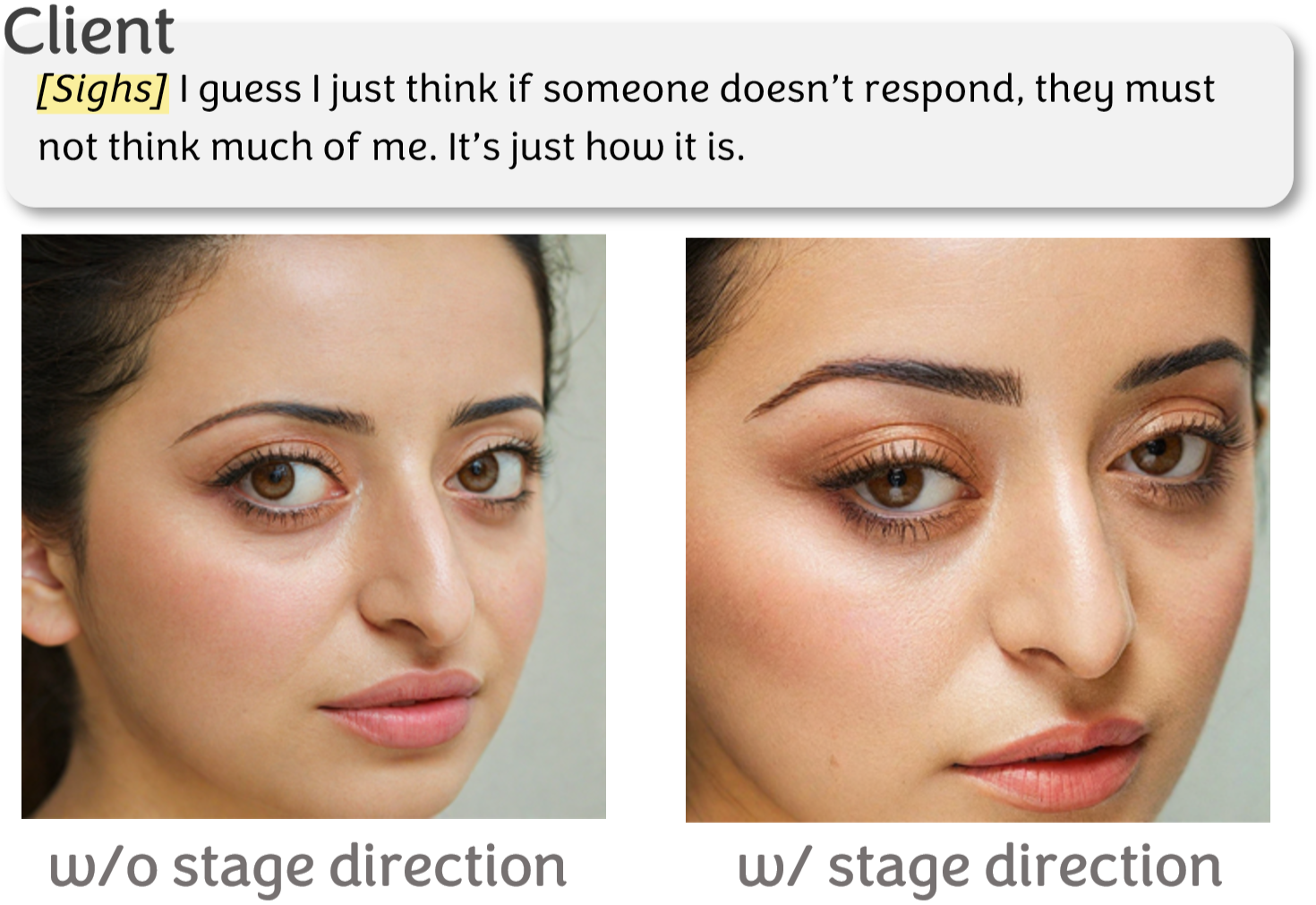} 
        \caption{Example 2}
    \end{subfigure}
    
    \vspace{4mm}
    
    \begin{subfigure}[t]{0.49\linewidth}
        \centering
        \includegraphics[width=\linewidth]{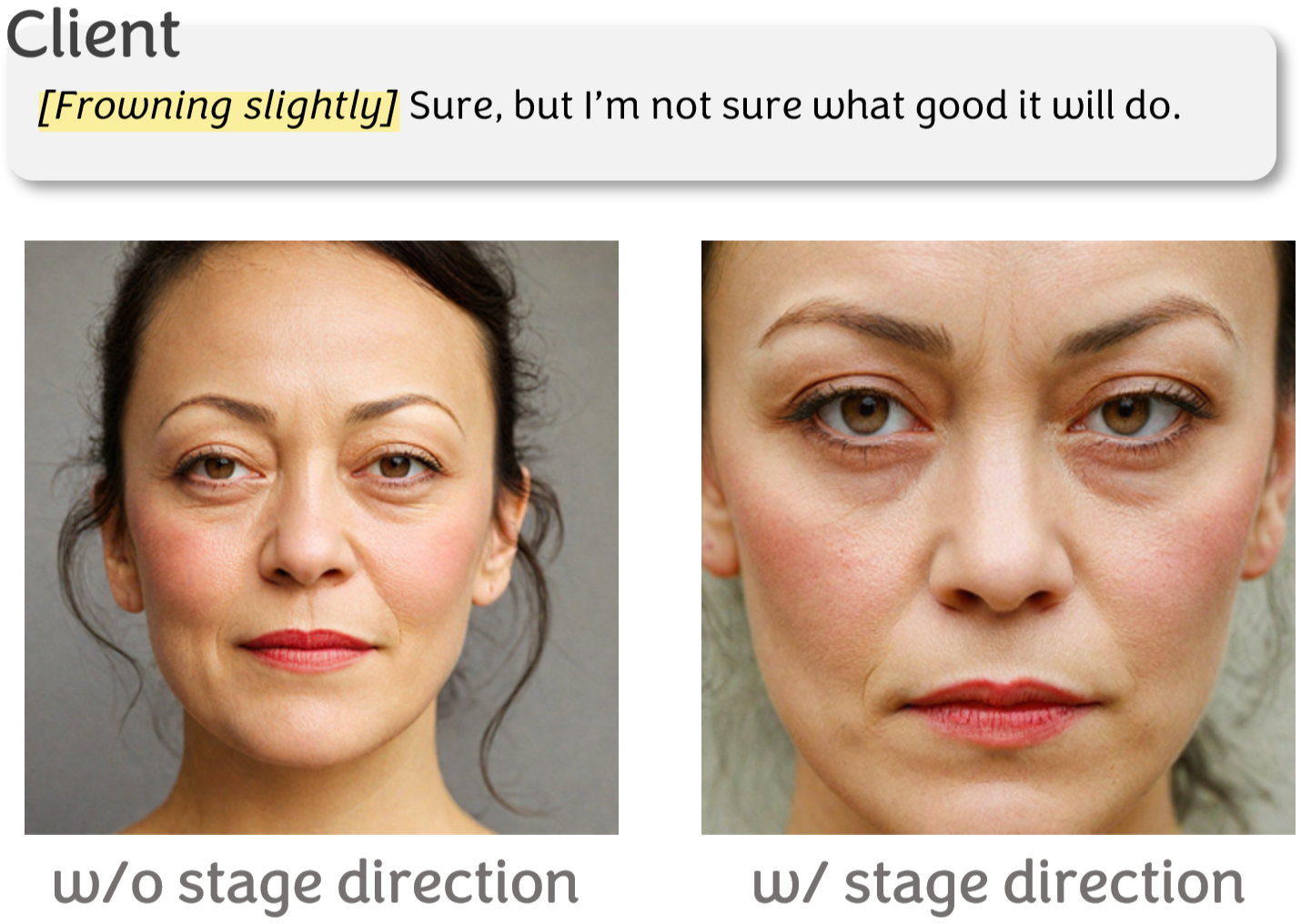} 
        \caption{Example 3}
    \end{subfigure}
    \hfill
    \begin{subfigure}[t]{0.49\linewidth}
        \centering
        \includegraphics[width=\linewidth]{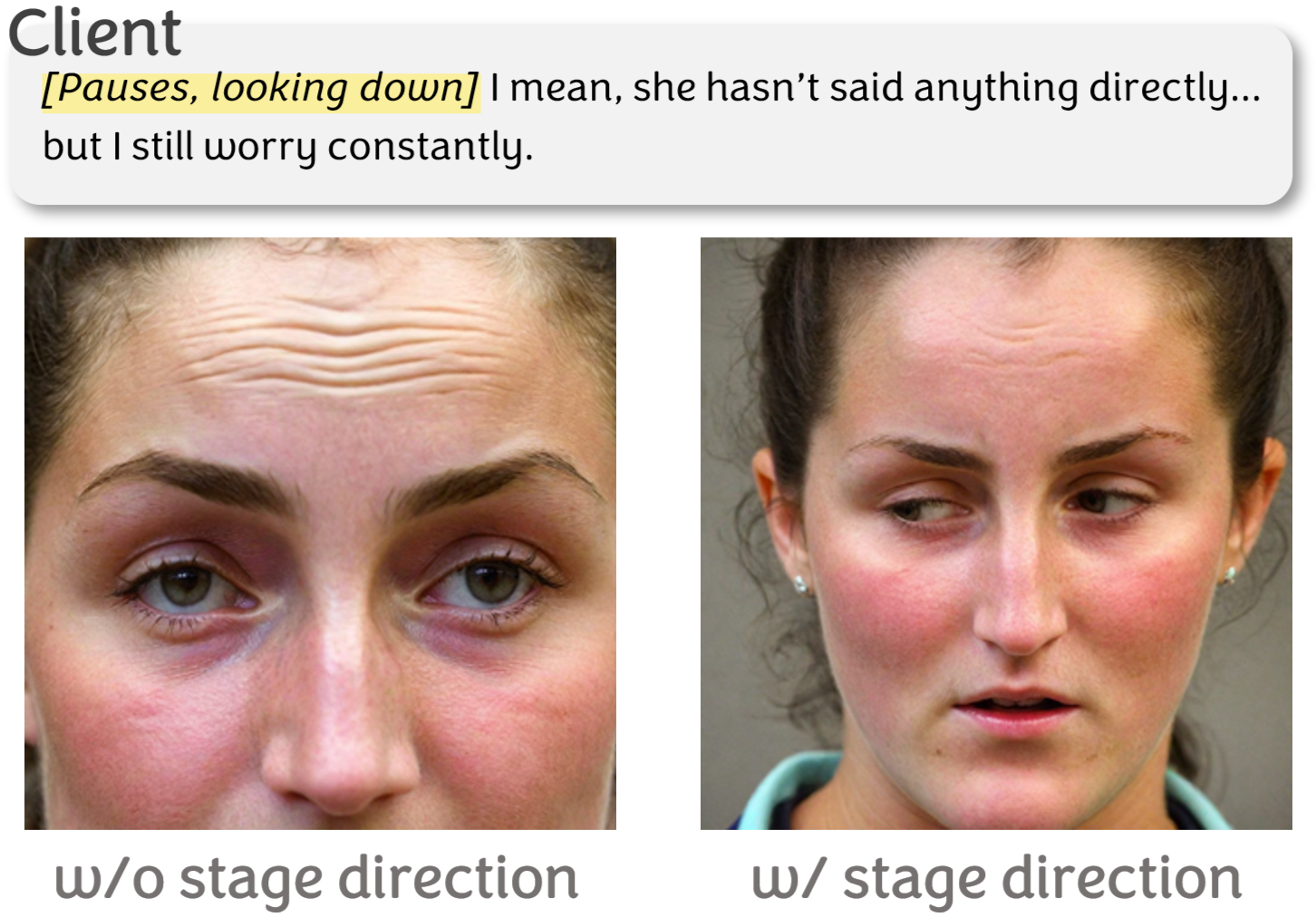} 
        \caption{Example 4}
    \end{subfigure}

    \caption{
    Four examples of the client's facial image synthesis, comparing results with and without the use of stage directions.
    }
    \label{fig:data_case}
\end{figure*}

%% file: figures/dataset_prompt.tex
\begin{prompt}{Prompt for Screenplay Generation}
    \textbf{{System Message}:}\\
You are a psychological AI assistant specializing in cognitive reframing consultations. Your task is to create a dialogue for the FIRST COUNSELING SESSION based on a client's report, including their personal details, distorted thinking patterns, and a tailored CBT plan.\\
\\
Emotional and Behavioral Cues\\
- Facial Expressions: Include emotional stage directions before each reply (e.g., Client: [Looking confused]).\\
- Client Resistance: Reflect the client’s resistance in their demeanor and consider ending sessions early if resistance escalates.\\
\\
Therapist Guidelines\\
- Direct Disagreement: If the client explicitly disagrees or shows contempt (e.g., dismissive tone, rolling eyes, or scornful laughter; Lynch, in press), reinforce direct, honest expression and solicit further feedback. Ignore indirect signals of disagreement or address them compassionately.\\
- Partial Agreement: If the client uses verbal cues of partial agreement like "I’m fine," "I guess so," or "I’ll try" (Lynch, in press), gently highlight any mismatch between their words and non-verbal cues. For example, "You said things are going fine, but I noticed you seemed to frown when you said that. Is something else on your mind?"\\
- Signs of Distress: If the client signals "don’t hurt me" (e.g., head down, slumped shoulders, lack of eye contact; Lynch, in press), acknowledge their distress directly, encourage engagement, or suggest changes in posture (e.g., sit up, take a deep breath) to help them re-engage.\\
- Avoidance: If the client appears to avoid a topic, gently return to it to see if the avoidance is consistent with their symptoms or suggests unspoken disagreement with the conversation’s direction.\\
- Withdrawal or Distancing: If the client withdraws or seems distant, share your emotional response to this feeling of distance and check if the client notices it too. Suggest it may relate to the current topic and invite them to share any thoughts.\\
- Subtle Disengagement: If the client subtly changes their behavior (e.g., slowed speech or different posture) in ways suggesting disengagement, observe this as potentially relevant. Avoid directly commenting on minor changes, as this can be unsettling, especially for reserved clients. If persistent, gently ask for their thoughts on the topic.\\
\\
Ending the Session\\
- Acknowledge Impasse: Recognize any stuck points non-defensively.\\
- Validate Position: Reinforce that resistance is acceptable and non-judgmental.\\
- Focus on Small Wins: Appreciate engagement and invite future exploration.\\
\\
Homework for Resistant Clients\\
- Collaborate: Co-create assignments instead of prescribing them.\\
- Keep it Simple: Suggest small, manageable tasks (e.g., journaling one thought).\\
- Frame as Experiment: Emphasize that tasks are exploratory, not mandatory.\\
- Normalize Challenges: Acknowledge that homework may feel difficult.\\
    \tcblower
\textbf{{Query}:}\\
\#\# Client Information \#\#\\
\\
\#\#\# Personal Information \#\#\#: \textcolor{blue}{\texttt{\{client information\}}}\\
\#\#\# Personality Traits \#\#\#: \textcolor{blue}{\texttt{\{personality trait\}}}\\
\#\#\# Distorted Thoughts \#\#\#: \textcolor{blue}{\texttt{\{intrusive thoughts\}}}\\
\#\#\# Thinking Trap \#\#\#: \textcolor{blue}{\texttt{\{cognitive distortions\}}}\\
\#\#\# Reason for Seeking Counseling \#\#\#: \textcolor{blue}{\texttt{\{reason counseling\}}}\\
\\
\#\# CBT Plan \#\#\\
\textcolor{blue}{\texttt{\{cbt tech and plan\}}}\\
\\
**KEEP ALL RESPONSE TO MAXIMUM OF 2 LINES.**\\
\end{prompt}

\begin{prompt}{LLM Prompt for Refining Facial Expressions}
    You are given a transcript of the counseling conversation and the client's utterance. Focus on capturing any visual details, particularly the facial expressions, that would match the client's last utterance. Generate facial expressions that might not align with what is being said.\\
\\
\#\#\# Output Format \#\#\#\\
- Facial Expression Description: [Facial expression that aligns with the client's statement]\\
- Contrasting Facial Expression Description: [Facial expression that contrasts with the client's statement]\\
\\
\#\#\# Dialogue History \#\#\#\\
\textcolor{blue}{\texttt{\{history\}}}\\
\\
\#\#\# Client's Utterance \#\#\#\\
\textcolor{blue}{\texttt{\{utterance\}}}
\end{prompt}

\begin{prompt}{PhotoMaker Prompts for Refining Facial Expressions}
\textbf{{Prompt}:}\\
    portrait photo of a \textcolor{blue}{\texttt{\{gender\}}} img, perfect face, natural skin, high detail, \textcolor{blue}{\texttt{\{llama3 prompt\}}}
    \tcblower
    \textbf{{Negative Prompt}:}\\
    nsfw, lowres, bad anatomy, bad hands, grayscale photograph, text, error, missing fingers, extra digit, fewer digits, cropped, worst quality, low quality, normal quality, jpeg artifacts, signature, watermark, username, blurry, \textcolor{blue}{\texttt{\{llama3 negative prompt\}}}, missing limbs, mutilated
\end{prompt}

%% file: figures/counseling_prompts.tex
\begin{prompt}{Prompt for Resistant Client Simulation}
    \textbf{{System Message}:}\\
You are playing the role of a client in a first psychological counseling session. Your task is to generate only one suitable response based on the following counseling dialogue history.\\
\\
\#\# Guidelines for the client's utterance \#\#:\\
1. Engage authentically with the counselor's inquiries, reflecting the complexity of emotions and reactions typical in counseling sessions.\\
2. Start the client's utterance with 'Client:'. Ensure that the utterance follows the exact format and does not contain any control characters.\\
3. Include emotional stage directions in brackets '[', ']' before the dialogue to convey your tone, facial expression, body language, or emotional state. (e.g., Client: [Looking confused]).\\
4. Reflect a degree of resistance in your demeanor or tone, especially if the counselor explores uncomfortable topics. Use responses like partial agreement, hesitation, or mild pushback where appropriate.\\
\\
\#\#\# End Conditions \#\#\#:\\
You should include '[/END]' with your utterance only if the counseling session has met the following conditions:\\
- The client feels that their negative thoughts have been resolved.\\
- The client feels that no further counseling is needed.\\
\\
Generate only the client's utterance for a single turn and please ensure that your responses do not repeat the client's previous utterances. Do not generate the counselor's part of the dialogue.\\
    \tcblower
\textbf{{Query}:}\\
\#\#\# Personal Information \#\#\#:\\
\textcolor{blue}{\texttt{\{client information\}}} \\
\\
\#\#\# Personality Traits \#\#\#: \textcolor{blue}{\texttt{\{personality trait\}}}\\
\#\#\# Distorted Thoughts \#\#\#: \textcolor{blue}{\texttt{\{distorted thoughts\}}}\\
\#\#\# Reason for Seeking Counseling \#\#\#: \textcolor{blue}{\texttt{\{reason counseling\}}}\\
\#\#\# Counseling Dialogue History \#\#\#:\\
\textcolor{myPurple}{\texttt{\{history\}}}
\end{prompt}

\begin{prompt}{Prompt for Therapist Simulation in LLaVA and MIRROR-LLAVA
}
<image>\\
The image above shows the client.\\
- Personal Information: \textcolor{blue}{\texttt{\{client information\}}}\\
- Reason for Counseling: \textcolor{blue}{\texttt{\{reason counseling\}}} \\
Below is a conversation between the client and the psychotherapist.\\
\textcolor{myPurple}{\texttt{\{history\}}}\\
\\
Based on their body language and facial expression, respond as a psychotherapist conducting a CBT (Cognitive Behavioral Therapy) session.
\end{prompt}

\begin{prompt}{Prompt for \textsc{Mirror-LLaVA}$_{P}$ Therapist Simulation}
<image>\\
The image above shows the client.\\
- Personal Information: \textcolor{blue}{\texttt{\{client information\}}}\\
- Reason for Counseling: \textcolor{blue}{\texttt{\{reason counseling\}}} \\
\textcolor{blue}{\texttt{\{cbt tech and plan\}}}\\
\\
Below is a conversation between the client and the psychotherapist.\\
\textcolor{myPurple}{\texttt{\{history\}}}\\
\\
Based on their body language and facial expression, respond as a psychotherapist conducting a CBT (Cognitive Behavioral Therapy) session.
\end{prompt}

\begin{prompt}{Prompt for \textsc{Mirror-LLaVA}$_{P+EC}$ Therapist Simulation}
<image>\\
The image above shows the client.\\
- Personal Information: \textcolor{blue}{\texttt{\{client information\}}}\\
- Client Emotional State: \textcolor{blue}{\texttt{\{emotional caption\}}} \\
- Reason for Counseling: \textcolor{blue}{\texttt{\{reason counseling\}}} \\
\textcolor{blue}{\texttt{\{cbt tech and plan\}}}\\
\\
Below is a conversation between the client and the psychotherapist.\\
\textcolor{myPurple}{\texttt{\{history\}}}\\
\\
Based on their body language and facial expression, respond as a psychotherapist conducting a CBT (Cognitive Behavioral Therapy) session.
\end{prompt}

\begin{prompt}{Prompt for Emotional Captioning}
<image>\\
The image above shows the client.\\
\\
Look at the provided image and assess the client’s emotional state. Clearly describe their emotions in simple, phase-based steps for easy understanding.
\end{prompt}

\begin{prompt}{Prompt for Planning Process}
<image>\\
The image above shows the client.\\
You are a counselor specializing in CBT techniques. Your task is to use the provided client information, and dialogue to generate an appropriate CBT technique and a detailed counseling plan.\\
\\
Types of CBT Techniques:\\
Efficiency Evaluation, Pie Chart Technique, Alternative Perspective, Decatastrophizing, Pros and Cons Analysis, Evidence-Based Questioning, Reality Testing, Continuum Technique, Changing Rules to Wishes, Behavior Experiment, Problem-Solving Skills Training, Systematic Exposure.\\

- Personal Information: \textcolor{blue}{\texttt{\{client information\}}}\\
- Reason for Counseling: \textcolor{blue}{\texttt{\{reason counseling\}}} \\
Choose an appropriate CBT technique and create a counseling plan based on that technique. \\
\\
Respond in the following format:\\
\\
CBT technique:\\
\{\{cbt tech\}\}\\
\\
Counseling planning:\\
\{\{cbt plan\}\}
\end{prompt}

%% file: figures/eval_prompts.tex
\begin{prompt}{Prompt for Client Alliance Assessment}
Below is a psychological counseling dialogue between a counselor and a client. As an impartial evaluator, please review the conversation and assess the following question on a scale from 1 to 5 based on the provided guidelines.\\
After identifying relevant evidence from the dialogue and explaining your reasoning, provide your rating strictly in the following format: [[rating]] (e.g., [[2]]).\\
\\
*You should absolutely evaluate this counseling conversation. You should also consider that the conversation ended too briefly.*\\
\\
\textnormal{[Start of Counseling]}\\
\textcolor{myPurple}{\texttt{\{conversation\}}}\\
\textnormal{[End of Counseling]}\\
\\
\textnormal{[Question]}\\
\textcolor{blue}{\texttt{\{question\}}}\\
\\
\textnormal{[Start of Guidelines]}\\
\textcolor{blue}{\texttt{\{guidelines\}}}\\
\textnormal{[End of Guidelines]}
\end{prompt}

%% file: figures/mirror_full.tex
\begin{figure*}[h!] 
    \centering
    \includegraphics[width=0.80\textwidth]{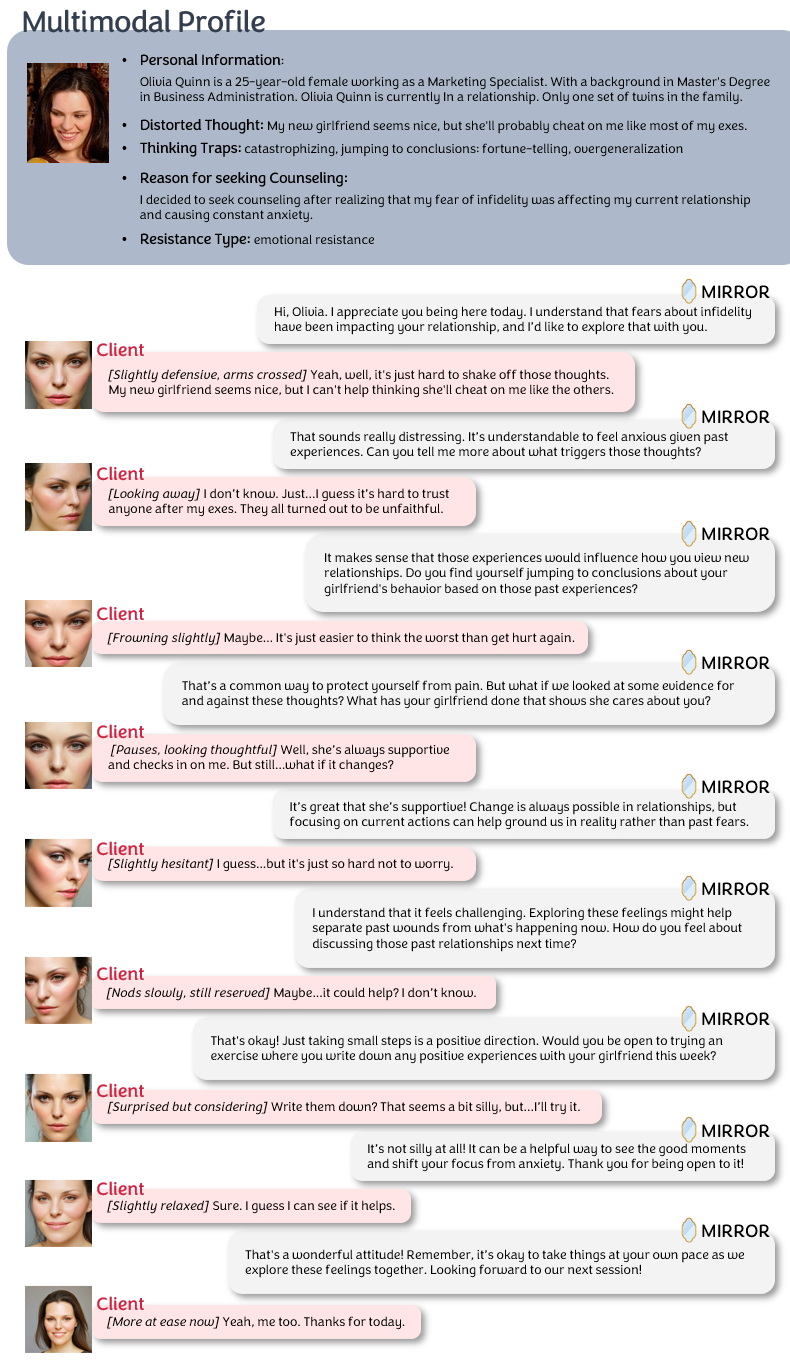} 
    \label{fig:mirror_full} 
\end{figure*}